\documentclass{article}

\usepackage{arxiv}

\usepackage[utf8]{inputenc}
\usepackage[T1]{fontenc}    
\usepackage{hyperref}       
\usepackage{url}            
\usepackage{booktabs}       
\usepackage{amsfonts}       
\usepackage{amsmath, amssymb}
\usepackage{nicefrac}       
\usepackage{microtype}      
\usepackage{lipsum}		
\usepackage{graphicx}
\usepackage[numbers]{natbib}
\usepackage{natbib}
\usepackage{doi}
\usepackage{xcolor}
\usepackage{enumitem}
\usepackage{bm}
\usepackage{multirow}
\usepackage{placeins}

\usepackage{color}

\newcommand\changes[1]{\textcolor{black}{#1}}

\title{Learning Mechanically Driven Emergent Behavior with Message Passing Neural Networks}

\author{{Peerasait Prachaseree}\\
	Department of Mechanical Engineering\\
	Boston University\\
	Boston, MA 02215 \\
	\texttt{pprachas@bu.edu} \\
	\And
	{Emma Lejeune} \\
	Department of Mechanical Engineering\\
	Boston University\\
	Boston, MA 02215\\
	\texttt{elejeune@bu.edu} \\
}

\date{}

\renewcommand{\shorttitle}{ABC Dataset and Metamodel}

\hypersetup{
pdftitle={Learning Mechanically Driven Emergent Behavior with Message Passing Neural Networks},
pdfsubject={mechanics, metamodels},
pdfauthor={Peerasait Prachaseree, Emma Lejeune},
pdfkeywords={Graph Neural Networks, buckling, classification, benchmark dataset},
}

\begin{document}
\maketitle

 \begin{abstract}
 From designing architected materials to connecting mechanical behavior across scales, computational modeling is a critical tool for understanding and predicting the mechanical response of deformable bodies. In particular, computational modeling is an invaluable tool for predicting global emergent phenomena, such as the onset of geometric instabilities, or heterogeneity induced symmetry breaking. Recently, there has been a growing interest in both using machine learning based computational models to learn mechanical behavior directly from experimental data, and using machine learning (ML) methods to reduce the computational cost of physics-based simulations. Notably, machine learning approaches that rely on Graph Neural Networks (GNNs) have recently been shown to effectively predict mechanical behavior in multiple examples of particle-based and mesh-based simulations. However, despite this initial promise, the performance of graph based methods have yet to be investigated on a myriad of solid mechanics problems. In this work, we examine the ability of neural message passing to predict a fundamental aspect of mechanically driven emergent behavior: the connection between a column's geometric structure and the direction that it buckles. To accomplish this, we introduce the Asymmetric Buckling Columns (ABC) dataset, a dataset comprised of three types of asymmetric and heterogeneous column geometries (sub-dataset 1, sub-dataset 2, and sub-dataset 3) where the goal is to classify the direction of symmetry breaking (left or right) under compression after the onset of the buckling instability. \changes{Notably, it is difficult to parameterize these structures into a feature vector for typical ML methods. Essentially, because the geometry of these columns is discontinuous and intricate, local geometric patterns will be distorted by the low-resolution ``image-like'' data representations that are required to implement convolutional neural network based metamodels.} Instead, we present a pipeline to learn global emergent properties while enforcing locality with message passing neural networks. Specifically, we take inspiration from point cloud based classification problems from the computer vision research field and use PointNet++ layers to perform classification on the ABC dataset. In addition to investigating GNN model architecture, we study the effect of different input data representation approaches, data augmentation, and combining multiple models as an ensemble. Overall, we were able to achieve good performance with this approach, ranging from $0.952$ prediction accuracy on sub-dataset 1, to $0.913$ prediction accuracy on sub-dataset 2, to $0.856$ prediction accuracy on sub-dataset 3 for training dataset sizes of $20,000$ points each. However, these results also clearly indicate that predicting solid mechanics based emergent behavior with these methods is non-trivial. Because both our model implementation and dataset are distributed under open-source licenses, we hope that future researchers can build on our work to create enhanced mechanics-specific machine learning methods. Furthermore, we also intend to provoke discussion around different methods for representing complex mechanical structures when applying machine learning to mechanics research.
 
\end{abstract}
  
\section{Introduction}
\label{sec:intro}

Advances in additive manufacturing have made it possible to create architected materials with unprecedented control and intricacy \citep{flexiblemetamaterials, holmesstability, biobone, resilientmeta, mesoprinting, nanoarchitect}. To design these materials, many researchers have turned to biologically-inspired design \citep{biobone, optimizationspiderweb}, where naturally occurring microstructure such as bones \citep{biobone}, and complex geometry such as spiderwebs \citep{optimizationspiderweb} are the starting point for achieving mechanical properties that exceeds what is possible with homogeneous domains and simple geometry. 
\changes{To complement this broad interest in exploring the design space of architected materials, multiple researchers have developed computational frameworks to both better understand and optimize architected material mechanical behavior, often through computationally expensive direct numerical simulations \citep{sanders2018multi, optimizemicrostructures}.
More recently, there has been a surge in applying machine learning (ML) methods to replace either computationally expensive simulations or resource demanding experiments with computationally cheap  metamodels, also known as surrogate models, that directly map input parameters to output quantities of interest (QoIs) \citep{reviewer2_3, reviewer2_1, reviewer2_2}.}
Within the broader context of ML based approaches for computational design and analysis frameworks, graph-based neural network approaches have received substantial recent attention for problems in recommendation systems \citep{gnnrec}, point cloud classification \citep{pointnet, edgeconv}, molecule property predictions \citep{gnnmolecule, mpnn}, and targeted drug design \citep{vae, GCPN}. 
Fundamentally, when it comes to designing complex structures, graph based techniques are appealing because they offer a natural strategy for representing domains that are not conveniently captured as ``image-like'' arrays. In the context of mechanics problems, this includes problems that involve structures such as trusses \citep{gnntruss}, disordered fiber networks \citep{amorphousnetwork}, and polycrystal microstructures \citep{gnnmicrostruc}. \changes{In this manuscript, our goal is to further explore the efficacy of these graph-based techniques for problems in solid mechanics, where global mechanically driven emergent behavior is highly influenced by local geometric properties.} 

Many examples of prior work on applying ML methods to complex structural design focus on structures that can be parameterized into a feature vector \citep{ bayesianmetamaterial,beartransfer, bear, spinodoid, BIC, leng2021predicting, optimizationspiderweb}. For domain information that is not conveniently represented as a low dimensional feature vector, treating domain geometry as ``image-like" arrays that serve as input to a Convolutional Neural Network (CNN) has become standard practice \citep{yang2020prediction}. For example, CNNs have been applied to predict change in strain energy \citep{mechmnist} and rank toughness from heterogeneous material distributions \citep{crackclass}, predict full-field mechanical QoIs \citep{DLfullfield, cGANfullfield}, aid in inverse design \citep{inversemembrane, inversekirigami}, and quantify uncertainty of partial differential equations (PDEs) \citep{bayesianfullfield}. For these and related examples, CNNs are an effective approach because they naturally enforce locality and spatial invariance through inductive bias with the convolution operation \citep{GN}. However, this strategy faces the fundamental limitation that arrays are often not an efficient approach for representing complex geometries, especially given the computational limitations associated with training a model on a dataset of large arrays \cite{cnnreview}.  \changes{In another recent approach, collocation methods have been used in conjunction with Physics-Informed Neural Networks (PINNs) where points are sampled within the domain as inputs to the neural networks \cite{PINNs, PINNbuckle}.} Alternatively, unordered and non-grid like data structures such as point clouds \citep{shapenet} and meshes \citep{faust} can be readily used to store geometric information for broader applications in both computer vision \citep{pointnet} and mechanical simulation \citep{gnnmesh}. Furthermore, there is broad interest in the ML research community on geometric learning, or extending ML methods to graphs and non-Euclidean data structures \citep{GN, geometriclearning}, where non-grid like datasets are represented as graph structures. Neural networks that directly take in graphs and operate on graph structures are called graph neural networks (GNNs), and with GNNs, locality and combinatorial generalization (i.e., constructing new inferences from known relations) can be induced through spatial graph convolutions \citep{GN}. 
In this work, we are interested in using GNNs to predict mechanically driven global emergent behavior from geometric structures. 

One of the most interesting aspects of solid mechanics is that components of a structure's local geometry and/or material properties often interact collectively to produce global emergent behavior such as symmetry breaking \citep{porebuckle}, auxetic behavior \citep{negativepoisson,porebuckle, amorphousnetwork}, and mechanical resilience \citep{resilientmeta}. Compellingly, these extreme behaviors often \textit{emerge} simply from the fundamental set of equations governing mechanical deformation \citep{holmesstability}. 
As data driven approaches to capturing mechanical behavior gain traction \citep{bayesianmetamaterial, beartransfer,  bear, inversekirigami, spinodoid, gansmetamaterial, optimizationspiderweb}, we anticipate that ensuring that data driven approaches are capable of capturing these emergent, and often highly non-linear, phenomena will be a key challenge. \changes{While there have been multiple successful examples of modeling buckling problems using data driven methods, these previous works do not explicitly incorporate the geometry of the structure into the ML model \citep{breadbuckle, PINNbuckle}.}
Critically, many recent advances in graph-based machine learning applied to solid mechanics problems have come from the computer science literature \citep{gnnmesh, gnnparticle}. In these initial examples, domain geometry has been relatively simple, and the problems have not necessarily exhibited compelling non-linearities. \changes{In this work, our goal is to not only investigate the efficacy of graph-based machine learning approaches applied to problems with emergent behavior, but also to define a test problem that other researchers can use to evaluate their own alternative frameworks.} 

To accomplish this goal, we first introduce the Asymmetric Buckling Columns (ABC) dataset that consists of columns with complex geometry generated from different probabilistic distributions. For every given column, the goal is to classify its buckling direction under compression as either ``left'' or ``right.'' Then, we investigate the efficacy of neural message-passing \citep{mpnn}, a GNN-based spatial convolution framework, for predicting this straightforward mechanically-driven global emergent behavior from local structure geometry. Specifically, we adapt the spatially-based graph convolution PointNet++ architecture originally developed to classify point cloud data structures to our dataset \citep{pointnet}. With the ABC dataset, we address the current lack of benchmark datasets for classification problems in mechanics. With the accompanying metamodeling pipeline, we introduce an approach to representing complex domain geometries as graph networks, and demonstrate the efficacy of spatial graph convolutions for making predictions on the resulting graph representations. Looking forward, we view this work as an important methodological step towards advancing computational design and analysis of architected materials where geometry plays an essential role in the behavior of the designed structure. 

The remainder of the paper is organized as follows. In Section \ref{Methods}, we first discuss the generation of our finite element analysis based ABC dataset. Then, we introduce our metamodeling pipeline that includes converting each domain geometry into a spatial graph and training a message-passing graph convolution based ML model. We end Section \ref{Methods} by briefly introducing simple ensemble methods to boost the performance of ML model prediction, \changes{as well as introducing methods to evaluate the efficacy of ensemble model confidence and uncertainty predictions}. Next, in Section \ref{Results}, we present the prediction accuracy \changes{as well as model confidence} of our framework, and highlight the most effective approaches to implementing our pipeline. Finally, we present concluding remarks and potential future directions in Section \ref{Conclusion}. We note briefly that information for accessing all associated data and code is given in Section \ref{additionalinfo}.

\section{Methods}
\label{Methods}

In this Section, we begin by introducing the open source Finite Element Analysis dataset that we generated and curated for this project -- the Asymmetric Buckling Columns (ABC) dataset. Then, we provide details of our graph neural network based metamodeling pipeline for approximating the results of these simulations. We note briefly that details for accessing the dataset, the associated simulation software, and the metamodel implementation are given in Section \ref{additionalinfo}. 

\subsection{Defining and Generating the ABC Dataset}
\label{sec:meth_dataset}

To date, most applications of ML methods to mechanics research have focused on problems in regression. In general, there has been less work within the mechanics community on classification problems. In addition, we note that many problems that have been treated as classification problems in mechanics could also be conveniently recast as regression problem with a threshold, e.g. tough vs. not tough \citep{crackclass}, stable vs. unstable \citep{BIC}. 
Critically, there are limited examples of open source benchmark datasets for testing the efficacy of ML pipelines in this area. 
To address this, we first introduce the Asymmetric Buckling Columns (ABC) dataset as a toy mechanics-specific classification dataset that is amenable to treatment as a graph structure and involves a global property that emerges from complex structure geometry. 
Specifically, the ABC dataset contains matched input and output pairs where each input is a column domain geometry and each output is the direction of symmetry breaking of the column under compression (either ``left'' or ``right''). The dataset consists of three sub-datasets with different algorithms for input domain generation. Details of domain geometry generation for all three sub-datasets are given in Section \ref{sec:meth_dgg}. Details of the Finite Element Analysis (FEA) simulations used to determine the direction of symmetry breaking for each geometry are given in Section \ref{sec:meth_fea}. Finally, we describe the data curation strategy and format in Section \ref{sec:meth_format}. 

\begin{figure}[ht]
        \centering
        \includegraphics[width=\textwidth,keepaspectratio]{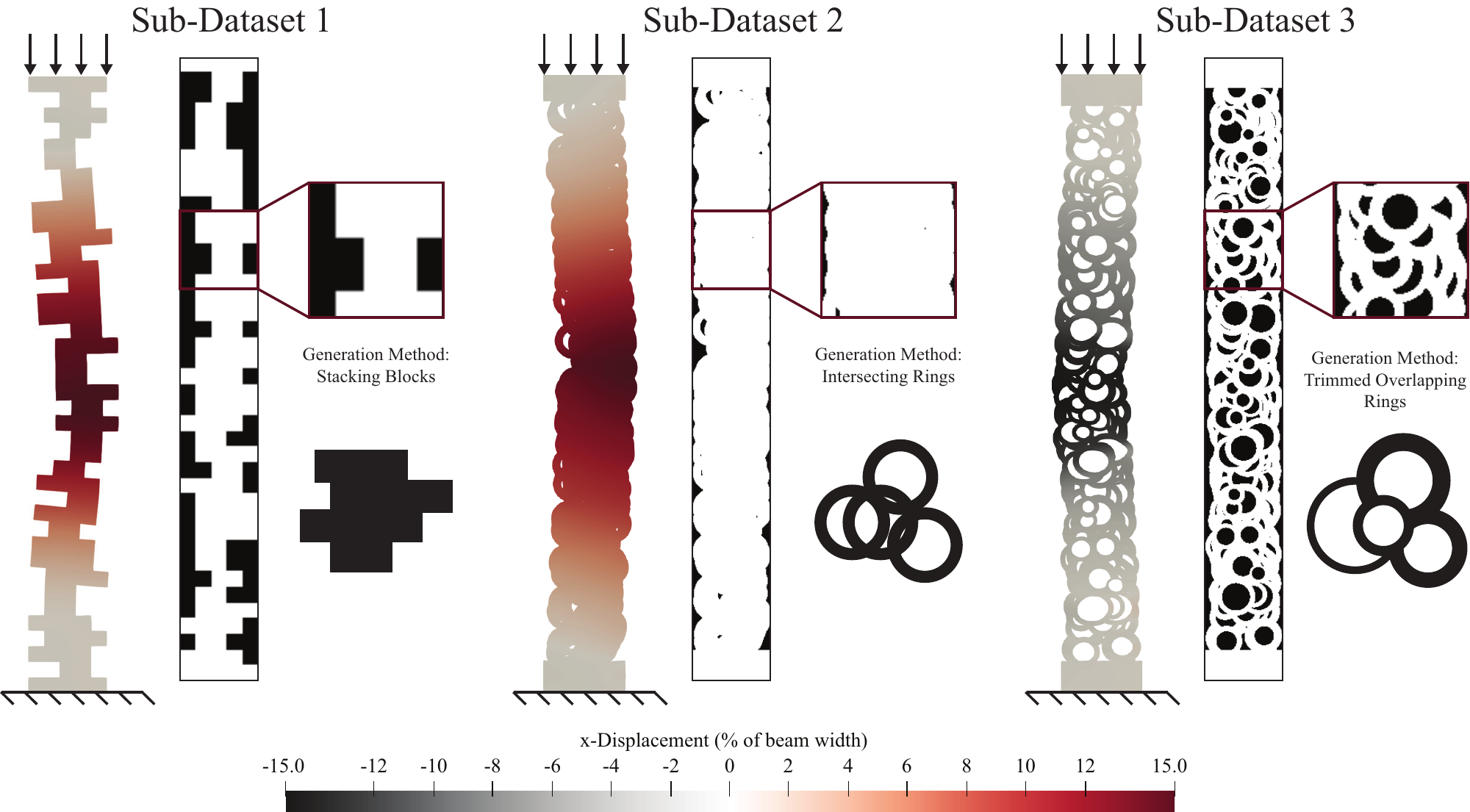}
        \caption{\label{fig:dataset} Schematic illustration of our Asymmetric Buckling Columns ``ABC'' dataset: The dataset consists of compressed inhomogeneous columns with fixed-fixed boundary conditions. The columns are compressed until the onset of buckling instability and the buckling direction is classified as either ``left'' ($0$) or ``right'' ($1$). The dataset is split into three sub-datasets, each with varying complexity of geometric features. The geometry of the first sub-dataset (sub-dataset 1) is generated by stacking rectangular blocks. The second sub-dataset is generated by intersecting rings of uniform size. The third sub-dataset is generated by overlapping and trimming rings of varying inner and outer radii. In this figure, deformed columns are visualized with ParaView \citep{paraview1,paraview2}.}
\end{figure}
        
\subsubsection{Domain Geometry Generation}
\label{sec:meth_dgg}

In Fig. \ref{fig:dataset}, we schematically illustrate all three sub-datasets of the ABC dataset. 
We chose to generate three sub-datasets to test if our metamodeling pipeline, described in Section \ref{sec:pipeline}, would function across multiple different input geometry distributions. 
For all three sub-datasets, the input domain generation begins with a rectangle with width ($w$) to length ($L$) ratio of 1:8. 
Then, all internal geometric features are algorithmically generated within the rectangular domain following procedures with varying geometric complexity. As illustrated in Fig. \ref{fig:dataset}, sub-dataset 1 has the simplest underlying features while sub-dataset 3 has the most complex underlying features. Sub-dataset 1 is generated by stacking $38$ blocks of height $0.025L$ with varying widths uniformly sampled from range $[0.4w,0.9w)$ (see Fig.\ref{fig:dataset} left). Note that the top-most and bottom-most blocks cover the whole width for all geometries for consistent boundary conditions. Sub-dataset 2 consists of $200$ to $300$ overlapping rings with outer radius of  $0.25w$ and inner radius of $0.15w$ (see Fig.\ref{fig:dataset} center). The centers of the rings are uniformly sampled from range $[0.25w,0.75w)$. Again, the top and bottom of the column have rectangular blocks of height $0.05L$ to enable the application of consistent boundary conditions across all structures. Sub-dataset 3 consists of $1000$ rings of different sizes that are overlapped and trimmed (see Fig.\ref{fig:dataset} right) The radii of the outer rings $R$ are uniformly sampled from range $R=[0.1w, 0.25w)$, and each inner ring radius $r$ is uniformly sampled from a range based on the corresponding outer ring radius defined as $r = [0.35R,0.75R)$. Again, the top and bottom of the column have rectangular blocks of height $0.05L$ to ensure consistent boundary conditions.

\subsubsection{FEA Simulations}
\label{sec:meth_fea}

We employ Finite Element Analysis (FEA) simulations to obtain the buckling direction of each column.
We use PyGmsh \citep{gmsh,PyGmsh} to mesh each domain, and open source software FEniCS \citep{fenics1,fenics2} to perform the FEA simulations. Specifically, we compressed the column under a displacement control loading protocol with the column fixed at the bottom and incrementally apply $y$ displacement at the top, schematically illustrated in Fig. \ref{fig:dataset}. The left and right side of each column is traction-free. We stop the FEA simulation once the magnitude of maximum $x$ displacement $u_x$ is greater than $0.15w$  ($\max |u_x| \geq 0.15w$). 
To determine the buckling direction, we compare the magnitude of the maximum and minimum $x$ displacements $u_x$. If $ | \max u_x | < |\min u_x |$, then the column is determined to buckle left and vice versa. We note briefly that we validated this approach by visualizing a random subset of FEA results with Paraview \citep{paraview1,paraview2}. 
For the purpose of ML model training, the columns that buckle left are assigned the label of ``$0$'' and the columns that buckle right are assigned the label of ``$1$.'' 

For all simulations, we used a compressible Neo-Hookean material model with the following strain energy density function:
\begin{equation}
    \psi = \frac{\mu}{2} \left[ \textbf{F}:\textbf{F}-3-2\ln{J} \right] + \frac{\lambda}{2} \left[ \frac{1}{2}(J^2-1) - \ln{J} \right]
\end{equation}
where $\psi$ is the strain energy density, $\textbf{F}$ is the deformation gradient, $J = \det{\textbf{F}}$, $\lambda$ and $\mu$ are the first and second Lam\'e parameters, respectively. The Lam\'e parameters are functions of Young's Modulus $E$ and Poisson's Ratio $\nu$ expressed as $ \lambda = {E\nu} / {[(1+\nu)(1-2\nu)]}$, and $\mu = {E}/{[2(1+\nu)]}$. 
For these simulations, we arbitrarily set the material parameters to $E=1$ and $\nu=0.3$. 
Prior to finalizing our simulation framework, we performed convergence studies to ensure that the direction of symmetry breaking was not sensitive to either the magnitude of incrementally applied loading or level of mesh refinement. We found that an incremental displacement of $(2.5\times10^{-4})L$ was sufficient for solution convergence with respect to loading \citep{javili2015computational,lejeune2016algorithmic,lejeune2016understanding}. We also found that a mesh of quadratic triangular elements with $\approx 1,300$ elements for sub-dataset 1, $\approx 27,300$ elements for sub-dataset 2, and $\approx 40,700$ elements for sub-dataset 3 was sufficient for solution convergence with respect to mesh size. 
We note briefly that in some cases the generated columns did not exhibit left-right symmetry breaking and instead either skipped to higher buckling modes or were dominated by local buckling behavior in a non-straightforward manner. This occurred in under $0.6\%$ of simulations across all sub-datasets. For simplicity, and to keep the focus on this manuscript on our ML modeling approach, we discarded these designs and generated new designs to obtain a total of $25,000$ structures with a clear direction of symmetry breaking per sub-dataset. For all sub-datasets, the direction of symmetry breaking is evenly split between the ``left'' and ``right'' classes.

\subsubsection{Note on data format}
\label{sec:meth_format}

As stated in Section \ref{sec:intro}, we designed the ABC dataset to address three needs: (1) the need for true classification mechanical benchmark datasets, (2) the need for mechanical benchmark datasets with non-grid like geometries, and (3) the need for additional mechanical benchmark datasets where global mechanical behavior emerges from local geometry. 
Releasing the ABC dataset as an open source mechanics-based classification dataset is a major contribution of this work. Here, we note briefly that we have made the ABC dataset inputs (i.e., the structure of each column) available through two independent approaches. First, in order to give other researchers maximum flexibility in adapting the components of our proposed pipeline, we provide the details necessary to reconstruct the exact input geometry. To accompany this information, we provide the code needed to re-generate the exact structures from these input parameters. Second, we directly provide graphs consisting of ``spare,'' ``medium,'' and ``dense'' nodal densities  generated with the pipeline proposed in Section \ref{sec:meth_representation} for each input geometry. This second approach is designed to make it as easy as possible for other researchers to begin working with the ABC dataset in PyTorch Geometric \citep{PyG}, a GNN library built on top of the popular ML framework Pytorch \citep{Pytorch}. Looking forward, we hope that other researchers will be able to use the ABC dataset to both design improved GNN architectures and devise new strategies to effectively represent the complex geometry of these structures. 
 Information for accessing the ABC dataset, and the accompanying code for both geometry reconstruction and importing graph structures into PyTorch Geometric, is given in Section \ref{additionalinfo}.

\subsection{Metamodeling Pipeline}
\label{sec:pipeline}

Our goal is to design a metamodeling pipeline that trains a ML model to effectively classify the direction of buckling (left vs. right) for the structures in our ABC dataset introduced in Section \ref{sec:meth_dataset}. We begin in Section \ref{sec:meth_representation} by outlining how we converted our heterogeneous columns into spatial graph network. The overall ML model architecture is introduced in Section \ref{ML}, and further details of our message-passing convolution layer implementation are given in Section \ref{graphconv}. We note that we employ the same set of ML model architecture and hyperparameters to separately train the metamodels for each sub-dataset. 

\begin{figure}[ht]
        	\centering
        	\includegraphics[width=\textwidth,keepaspectratio]{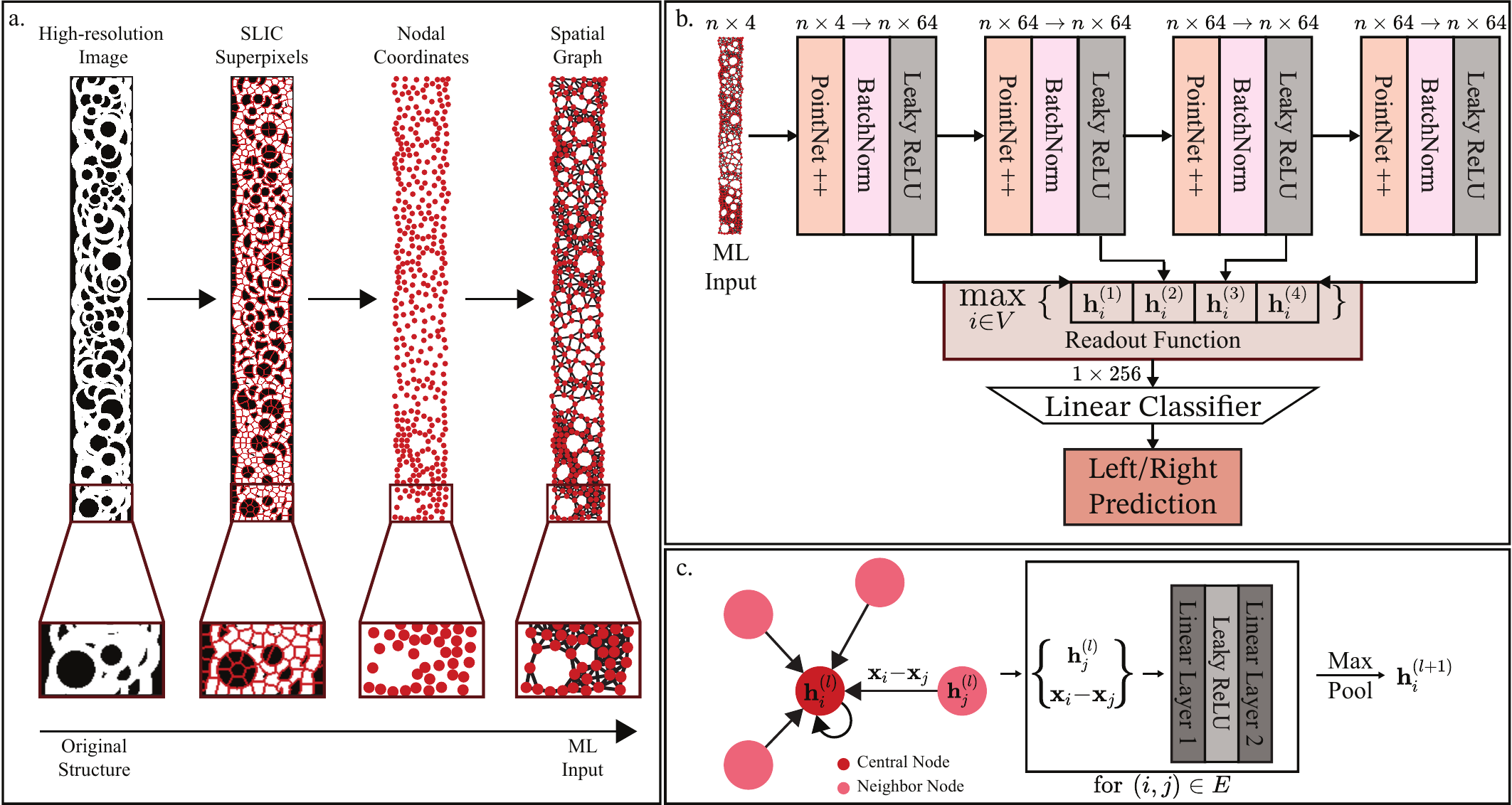}
        	\caption{An overview of our metamodeling pipeline: a) Steps to convert the high-resolution image representation of the structure to an undirected graph. b) Architecture used to to predict left/right symmetry breaking. We use $4$ PointNet++ Layers to generate embeddings  $\textbf{h}_i^{(l)}$ (nodal features of node $i$ in layer $l$) \citep{pointnet} with skip-connections \citep{resnet} after each batch normalization layer to obtain the final node embedding, and global max-pooling as the readout function before passing the final embedded vector through a linear classifier for ``left'' vs. ``right'' prediction. The corresponding labels $n \times 4$ and $n\times 64$ denote the number of nodes $n$ and the length of the current embedding vector. c) A schematic of our implementation of a Pointnet++ Layer with spatial graph convolutions. We construct our message functions for message-passing by concatenating node features $\textbf{h}_j^{(l)}$, which are the embedded feature vector of the neighboring node, and edge features, which are formed from the difference of positions $\textbf{x}_j-\textbf{x}_i$. The message function $\textbf{h}_i^{(l)}$ is updated using a Multi-Layer Perceptron (MLP) with one hidden layer aggregated with max pooling. The resulting vector $\textbf{h}_i^{(l+1)}$ is the new embedding that is passed in the next step of the architecture.}
        	\label{fig:ML}
        \end{figure}

\subsubsection{Representation of Structures}
\label{sec:meth_representation}

As stated in Section \ref{sec:intro}, the main focus of this work is predicting the direction of buckling deformation for complex geometric structures. Specifically, we are interested in structures where representing them coarsely as image-like arrays would be inefficient and/or potentially distort their geometry and lead to loss of critical information.  
In this work, we will represent these complex input data with minimal loss of information by converting each geometric structure into an undirected graph \citep{graph}. As a brief background, graphs are ordered pairs $G = (V,E)$ consisting of $n$ nodes (also referred to as vertices) $V = \{v_1, v_2, \dots, v_n\}$  and edges $E \subseteq V \times V$ (edges are constructed by connecting nodes). For undirected graphs, the edges are bidirectional. In general, graphs can contain both nodal features $\textbf{f}_i$ that contains information associated with nodes, and edge features $\textbf{e}_{ij}$ that define the relationship between two nodes. 

In Fig. \ref{fig:ML}a, we show our methodology for converting geometric structures into undirected graphs. We note that every step in our pipeline is modular, and other methods to generate nodes and determine their connectivity, such as methods in point cloud processing \citep{edgeconv}, could also be applied in the context of our broader analysis pipeline. The first step of our approach is to create a high resolution image representation ($100\times800$) of the domain. Then, we segmented the high resolution image array with simple linear iterative clustering (SLIC) \citep{SLIC} using the package sci-kit image (skimage) \citep{skimage}. With this approach, each segmented ``superpixel'' becomes a node in a spatial graph with nodal features of centroid position, area, and eccentricity. Since the SLIC algorithm can vary the number of superpixels in each image, it is possible to get a different density of nodes for the same structure by changing this parameter. The nodal density of the graph is a parameter that we will investigate in Section \ref{sec:res_rep}. 
After segmentation, the Region Adjacency Graph (RAG) of the structure is formed by discarding superpixels not associated with the structure (i.e., the superpixels associated with the dark areas in Fig. \ref{fig:dataset}), and adding edges between adjacent remaining superpixels. 
The RAG is one method of data representation that we will explore in Section \ref{sec:res_rag}. 
In addition, we also investigate a ``Ball Query'' based method of data representation where the structure-associated superpixel nodes are connected to form an undirected graph via a ball query algorithm \citep{pointnet}. Specifically, the ball query algorithm constructs edges by searching for nodes within a prescribed and tunable radius from the central node ($ E_{ij} = \{\{i,j\}  \; | \; \| \textbf{x}_j - \textbf{x}_i \| \leq r \; \textrm{and} \; i,j \in V\} $) where $\textbf{x}_i$ and $\textbf{x}_j$ are the positions of the central and neighboring node respectively and $r$ is the radius of ball query). For our spatial graphs, the edges for both the RAG and ball query approaches carry feature vectors $\textbf{x}_j-\textbf{x}_i$ that describe the distances between the two nodes. We note that all the node feature vectors are normalized after ball query to ensure faster training convergence.

Alternatively, we note that structures from sub-dataset 1 and sub-dataset 2 can also be simply represented as spatial graphs without loss of information. In Section \ref{sec:res_rag}, we investigate the effectiveness of using these ``exact structures'' as ML model inputs. In the case of sub-dataset 1, we obtain ``exact structures'' by first converting each pixel of the structure into a node with a nodal feature vector containing only nodal position. Then, we connect neighboring pixels with edges that contain the relative difference in positions of connected nodes ($\textbf{x}_j-\textbf{x}_i$). Since each pixel is the same size and shape, additional information regarding the geometry of each pixel would be redundant. For sub-dataset 2, we represent the ``exact structure'' by treating the center of each ring as a node where graph edges denote rings that overlap each other. Similarly, the edges contain the relative difference in positions of the connected nodes ($\textbf{x}_j-\textbf{x}_i$). Again, since the shape and size of each ring in sub-dataset 2 is identical, additional information regarding shape and size would be redundant. With this approach, no information is lost in describing the geometry -- thus we refer to it as ``perfect representation.'' We note briefly that for sub-dataset 3, the trimmed rings (see Fig. \ref{fig:dataset}) preclude a perfect representation approach. 

\subsubsection{Machine Learning Model Architecture} 
\label{ML}
The goal of our ML task is to predict the direction of buckling (left or right) from the geometry of the structure. Since buckling behavior is a global emergent behavior that arises from local geometric features \citep{flexiblemetamaterials, holmesstability, poreshapebuckling}, we chose to use a convolution based approach which inherently enforces locality through inductive bias. However, traditional Convolutional Neural Networks (CNNs) \citep{cnn} are not designed for data that is not grid-like (e.g., graph structure data \citep{graph}, point cloud data \citep{shapenet}, mesh data \citep{gnnmesh}). Instead of CNNs, we draw from recent interest within the computer vision research community in classifying point clouds, unordered and non-gridlike sets of points in space, with Graph Neural Networks (GNNs) and spatial graph convolutions \citep{MoNet,edgeconv}. These geometry-encoding point clouds are related to our geometric graph structures because they are both required to encode \textit{spatial} information to drive classification. As such, our ML model starting point is the previously developed PointNet++ architecture of spatial graph convolution layers \cite{pointnet}. While one of the perceived drawback of PointNet++ is the lack of inherent rotation invariance in the convolution layers \citep{pointnet}, not having inherent rotation invariance is advantageous for our dataset where we classify left-right symmetry breaking since the direction of buckling (and the corresponding labels) change depending on the orientation of the column. In Fig. \ref{fig:ML}b, we schematically illustrate our proposed machine learning model architecture. Specifically, we applied $4$ PointNet++ layers (See section \ref{graphconv}) with skip connections \citep{resnet} to help smooth the gradient flow during back propagation. In the context of graph convolutions, skip connections have also been shown to help prevent over smoothing by preserving local information \citep{gnn}. We implement these skip connections by concatenating each embedded nodal feature vector ($\textbf{h}_i^{l}$) after every convolution layer into a final nodal embedding vector (see Fig \ref{fig:ML}b). The global graph embedding is then obtained by using global max-pooling on the concatenated nodal features as the readout function (highlighted $\max_{i \in V}$ in Fig. \ref{fig:ML}b). The resulting global graph embedding vector is subsequently used to predict the direction of symmetry breaking via a linear classifier. To train this model, we use the cross-entropy loss function, and the ML model is trained for $50$ epochs with an Adam optimizer \citep{adam} before convergence. In short, the ML algorithm takes in nodal features and computes the global graph embedding through spatial graph convolution layers. The global graph embedding is then passed through a linear classifier to predict the buckling direction of the structure. To ensure reproducibility, all ML models are trained with $10$ different weight initialization seeds. For each sub-dataset,  the ML model is trained on $20,000$ data points. We also initially deployed and tuned the hyperparameters and architecture of our ML models on validation data ($2,500$ data points) before finally evaluating our model on held out test data ($2,500$ data points). 

\subsubsection{Spatial Graph Convolution Layer Implementation}
\label{graphconv}

We implemented the PointNet++ spatial graph convolutions using the Pytorch Geometric library \citep{PyG}. For tasks involving prediction of whole graphs, GNNs that utilize spatial graph convolutions, also known as message passing neural networks \citep{mpnn}, can be summarized in two phases: the message passing phase, and the readout phase. During the message passing phase, the nodal embeddings are recurrently updated through ``messages.'' These ``messages'' are constructed from current central nodal embeddings $\textbf{h}_i^{(l)}$ (or nodal feature vectors $\textbf{f}_i$) and neighboring embeddings $\textbf{h}_j^{(l)}$. Note that neighboring nodes in terms of graphs refer to nodes that are directly connected by an edge (see Fig\ref{fig:ML}c). New nodal embeddings $\textbf{h}_i^{(l+1)}$ are obtained by putting the current embeddings through an update function and combining them by an aggregation function. While the update function can be any differentiable function \cite{GN,mpnn}, in the case of GNNs, the most common update function is a Multilayer Perceptron (MLP) \citep{mlp}. The aggregation function is also a differentiable function with the additional constraints of being invariant to input node order and able to take in a variable number of inputs. Examples of commonly used aggregation function include element-wise sum, mean pooling, and maximum/minimum pooling \cite{GN}. Each update-aggregation pairing is also referred to as a spatial graph convolution layer. Note that in general, while our implementation passes the message through the update function before the aggregation function, the order of update and aggregation can be interchanged \cite{GN}. During the message passing phase, the nodal embeddings will be updated a prescribed number of times prior to the readout phase. During the readout phase, we then compute a global graph feature vector through readout functions that are also invariant to node ordering. Common readout functions, similar to aggregation functions, are mean-pooling, min/max-pooling, and sum-pooling applied to all of the node embeddings in the graph \citep{GN}. For ease of implementation, we employ EdgeConv \citep{edgeconv} (a generalized framework for performing spatial convolutions on point clouds) to perform spatial graph convolutions during the message passing phase on our spatial graph. In general, EdgeConv also uses the edge features alongside nodal features to construct messages for message passing. We note again briefly that since our pipleline is modular, variants of EdgeConv or alternative spatial graph convolutions layers could be applied here instead in future work.

For our ML model implementation of message passing, each node feature $\textbf{f}_i$ consists of nodal position $\textbf{x}_i$, superpixel area, and superpixel eccentricity (eccentrity is defined as $e=\frac{c}{a}$ where $c$ and $a$ are the focal distance and major axis length respectively of an ellipse with the same second moment of inertia as the superpixel). Each edge feature contains the difference in position between the central node and the connected neighboring node ($\textbf{x}_j-\textbf{x}_i$ with $\textbf{x}_i$ and $\textbf{x}_j$ denoting the position of the central and neighboring node respectively). As seen in Fig. \ref{fig:ML}c, each Pointnet++ layer constructs its message by concatenating the current nodal features and the edge features. The message is passed through a MLP \citep{mlp} with $1$ hidden layer (our update function), and aggregated with max-pooling to update the nodal embedding. Note that self-loops (edges that connect the central node to itself) are present in our graphs. As such, the new embedded feature vector $\textbf{h}_i^{(l+1)}$ is expressed as:

\begin{equation}
    \textbf{h}_i^{(l+1)} = \max_{(i,j) \in E} \text{MLP}\{\textbf{h}_j^{(l)}, \textbf{x}_j-\textbf{x}_i\}
\end{equation}

where $\textbf{h}$ denotes the nodal embedding vectors. Subscripts $i$ and $j$ denote the central node and neighboring node respectively, and superscript $l$ denotes the convolution layer number.
We initialize $\textbf{h}_i^{(0)}$ as $\textbf{f}_i$ (i.e., the first input to the first PointNet++ Layer is the nodal feature vector). After the first layer, all the embedding vectors have a length of $64$. We note that all features except for position are min-max scaled. After every PointNet++ Layer, we apply batch normalization (BatchNorm) \citep{batchnorm} to accelerate training and to provide some additional regularization. We apply the Leaky ReLu activation function \citep{leakyrelu} after the BatchNorm layer, and the resulting embedding nodal vectors are passed onto the next layer. After the last convolutional layer, our readout function is max-pooling, as described in Section \ref{ML}. In summary, during the message passing phase, our spatial convolutional layers construct messages from initialized nodal features and edge features that recurrently update with each convolutional layer. For the readout phase, we use max-pooling to generate an embedding of each graph and then use it to classify the buckling direction of each corresponding geometric structure.

\subsubsection{Ensemble Learning}
\label{sec:meth_ensemble}
In this work, we train our ML model $10$ times for each dataset with $10$ different random weight initializations (i.e., different seeds). For each ML model initialization, we obtain a slightly different set of predictions. Thus, each ML model acts as a different classifier. With these different classifiers on hand, we are then able to use ensemble methods to increase the overall prediction accuracy of our ML model framework. Broadly speaking, there are many different ensemble methods such as boosting \citep{boosting}, stacking \citep{stacking}, and bagging \citep{bagging}, that have been shown to increase overall prediction accuracy by leveraging multiple ML models \citep{Cnnensemble,ensemblednn, ensemble}. In this work, we employ a simple voting approach that requires no additional training beyond creating the initial ML models \citep{Cnnensemble}. Specifically, we investigate if a voting based approach will enhance our GNN ML model framework. We investigate predictions obtained by combining $10$ trained ML models through both hard and soft voting.
For hard voting, also referred to as majority voting, the predicted labels for each input are counted and the label with the highest count becomes the final prediction. 
For soft voting, also referred to as unweighted model averaging, the predicted probability of classes for each label is averaged over all ML models. The final predicted label is then the class with the highest average probability. In Section \ref{sec:res_vote}, we show the performance of both hard and soft voting.
\changes{Additionally, since each individual ML model is trained with a cross entropy loss, the computed probabilities via soft voting are also the ensemble's prediction confidence \citep{deepensembleuncertainty}. Broadly speaking, ML models are most useful for engineering applications when they are able to produce a meaningful confidence level for each prediction. In this context, a ML model is considered ``well-calibrated'' when model confidence reflects the true probability that the model is correct \citep{calibration, ECEMCE, reliability}. By way of example, if a ML model is perfectly calibrated and is $80\%$ confident for $100$ predictions, $80$ of those predictions will be correct. Reliability diagrams, also referred to as calibration curves, are a common visual approach to assessing the quality of model confidence. Briefly, reliability diagrams are constructed by first separating the prediction confidence that a given sample is in a chosen class into sequential bins, and then comparing the bin average to the true fraction of samples with the chosen class in each bin. In the context of the ABC dataset, we can construct a reliability curve by choosing the ``right'' buckling class and placing ``model confidence'' on the x-axis and ``fraction buckled right'' on the y-axis.  A perfectly-calibrated  model will exhibit a reliability curve equal to the identity function \citep{calibration, ECEMCE, reliability}. 
In addition to visualizing these curves, we also report quantitative metrics for model calibration. Specifically, we compute the Expected Calibration Error ($ECE$) and the Maximum Calibration Error ($MCE$) from each reliability diagram \cite{ECEMCE}. The $ECE$ is computed by taking a weighted average the of gaps between between perfect calibration and model confidence within each bin, written as:
\begin{equation}
    \label{eqn:ECE}
    ECE = \sum_{i=1}^B \frac{n_i}{N}|F_i-C_i|
\end{equation}
where $B$ is the number of bins, $n_i$ is the number of samples in each bin, $N$ is the total number of samples, $F_i$ is the frequency of class $1$ in the bin, and $C_i$ is average confidence that the sample is in class $1$ in the bin). 
The $MCE$ is defined as the maximum gap in each reliability diagram, written as:
\begin{equation}
    \label{eqn:MCE}
    MCE = \max_{i \in \{1, \dots B \}}|F_i-C_i| \, . 
\end{equation}
In Section \ref{sec:calibration}, we investigate the model calibration of our best-performing ensembles in each sub-dataset. }

\section{Results and Discussion}
\label{Results}

In this Section, we investigate multiple approaches towards designing a metamodeling pipeline and subsequently increasing its prediction accuracy. Section \ref{sec:res_aug} describes our data augmentation strategy that takes advantage of the symmetries in our training dataset. Next, in Section \ref{sec:res_rep}, we explore the effect of graph nodal and edge density on the prediction accuracy. Then, in Section \ref{sec:res_rag}, we perform a brief study on different strategies to represent structures as graphs. \changes{In Section \ref{sec:res_vote} we leverage ensemble learning to obtain higher prediction accuracies. We end by examining model calibration in Section \ref{sec:calibration}} We note that all results reported from here onward are obtained from held out test data.

\subsection{Data Augmentation Enhances Prediction Accuracy}
\label{sec:res_aug}

\begin{figure}[ht]
        	\centering
        	\includegraphics[width=\textwidth,keepaspectratio]{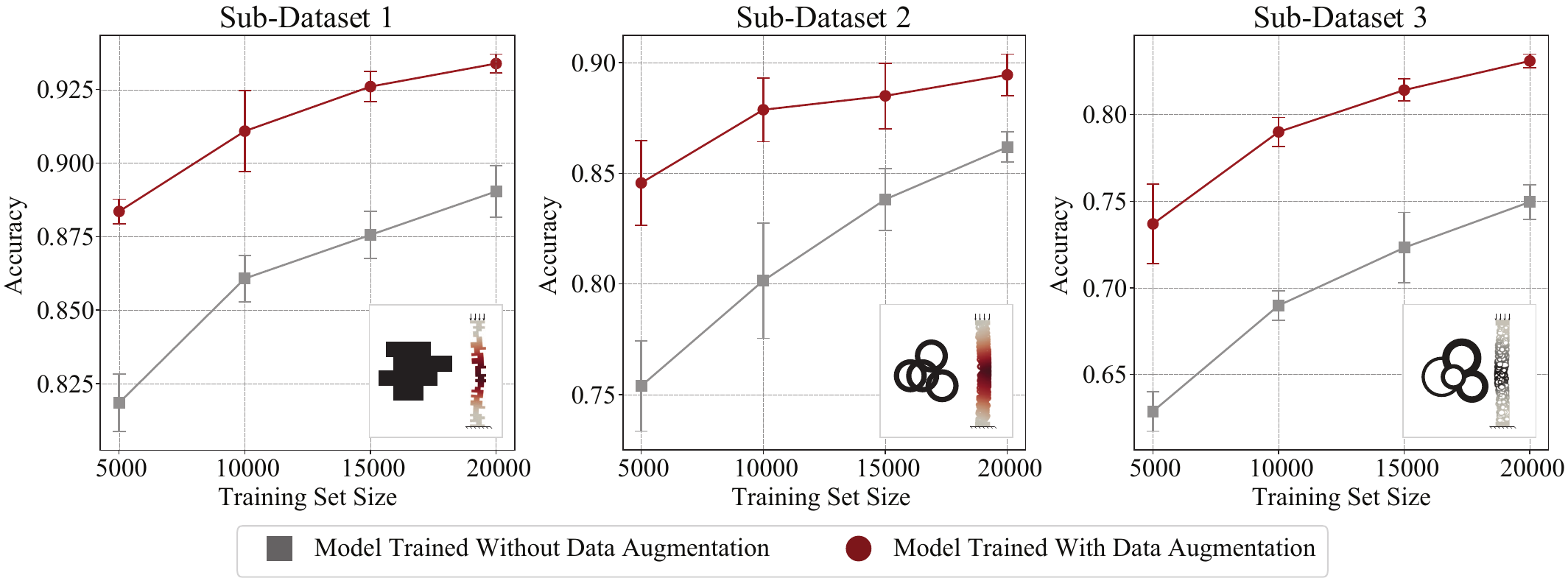}
        	\caption{Effect of data augmentation on test accuracy. The markers show the mean of $10$ trained models with different weight initialization and the error bars represent a $95$\% confidence interval. Note that the scales for the y-axis are different for each plot. The dataset is augmented by reflecting the columns about the $x$ axis, $y$ axis and both axes. Note that the output labels is changed when the column is reflected about the $y$ axis and both axes.}
        	\label{fig:aug}
        \end{figure}
        
Here, we present the performance of our PointNet++ architecture on both un-augmented and augmented versions of each sub-dataset. 
We perform data augmentation by reflecting each column in the input domain about the $x$ axis, $y$ axis and both axes. When the column is reflected about the $y$ axis and both axes, the associated output label is changed due to the symmetry of the problem. This method of data augmentation is only feasible because of the lack of rotational invariance in the PointNet++ Layers.
We note briefly that in these studies, we begin data augmentation with the ground truth geometries of the columns before segmentation and conversion into point clouds (see Fig. \ref{fig:ML}a). The augmentations are then combined with the original training dataset. By reflecting the geometries rather than the graphs themselves, the graphs of the reflected structure will be slightly different than if the graphs in the original training dataset had simply been rotated due to the randomness associated with the segmentation process. These variable graph representations also enforce the notion that slightly different graphs can represent the same structure.

\begin{table}[ht]
    \centering
    \caption{Comparison of the model test accuracy on the original training dataset and the augmented training dataset. Tabulated values are the mean and $95\%$ CI of the model with 10 different initializations. The values shown in this table are the best prediction accuracies we obtained using this metamodeling pipeline. Prediction accuracies are obtained from Fig. \ref{fig:aug}. }
    \begin{tabular}{|l||cc|cc|}
    \hline
    & \multicolumn{2}{|c|}{\textbf{Original Dataset}} & \multicolumn{2}{|c|}{\textbf{Augmented Dataset}} \\
    & \small{$5$k training points} & \small{$20$k training points} & \small{$5$k training datapoints} & \small{$20$k training datapoints}\\
    \hline
         \textbf{Sub-Dataset 1} & $0.818\pm0.010$ & $0.890\pm0.009$ &  $0.884\pm0.004$ & $0.934\pm0.003$\\
         \textbf{Sub-Dataset 2} & $0.754\pm0.020$ & $0.861\pm0.007$ &  $0.846\pm0.019$ & $0.894\pm0.009$\\
         \textbf{Sub-Dataset 3} & $0.628\pm0.011$ & $0.749\pm0.010$ &  $0.737\pm0.023$ & $0.831\pm0.004$\\
    \hline
    \end{tabular}
    \label{tab:aug}
\end{table}

In Fig.\ref{fig:aug} and Table \ref{tab:aug}, we demonstrate the effectiveness of data augmentation in increasing model predictive performance. Note that each data point contains the mean $\pm 95\%$ confidence interval (CI) of $10$ separately trained models with the same architecture but with different weight initializations.
In Fig. \ref{fig:aug}, we show test prediction accuracy with respect to the number of unaugmented training datapoints for both unaugmented and augmented training datasets. 
The results shown in Table \ref{tab:aug} (values obtained from Fig. \ref{fig:aug}) are the highest prediction accuracy for each sub-dataset.
Based on these results, it is clear that our approach to data augmentation is highly effective at increasing prediction accuracy. Specifically, it increases prediction accuracy by $~3\%$ to $~8\%$ in the case of $20,000$ training points. 
And, from the results presented in Table \ref{tab:aug}, we can see that the accuracy between training the model without data augmentation with $20,000$ data points (Table \ref{tab:aug} column $1$) is nearly equivalent to training the model with $5,000$ data points (Table \ref{tab:aug} column $2$) but with data augmentation. This demonstrates that our data augmentation method is comparable to increasing the training set size by a factor of $4$.
In addition, as shown in Fig. \ref{fig:aug}, even with data augmentation, the trend of the accuracy is still increasing at $20,000$ datapoints. This indicates that more datapoints (i.e., a larger training dataset) would likely improve the model prediction accuracy. 
We also note that as anticipated, the highest prediction accuracy is in sub-dataset 1, which has the simplest features, while sub-dataset 3 has the lowest accuracy corresponding to the most complicated features.

\subsection{Node Density and Query Radius Influence Prediction Accuracy}
\label{sec:res_rep}
\begin{figure}[ht]
        	\centering
        	\includegraphics[width=\textwidth,keepaspectratio]{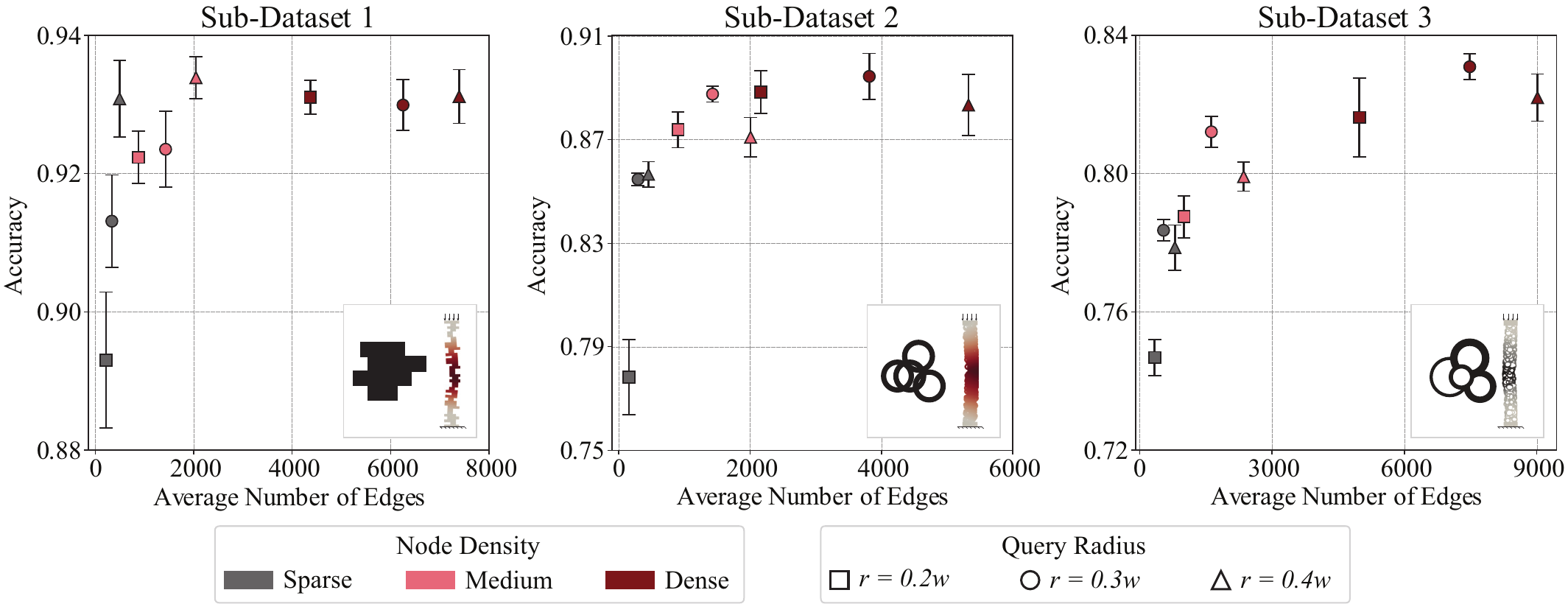}
        	\caption{These plots show the importance of graph structure representation. Each marker represents the mean of 10 trained models with different initializations and the error bars represent the $95\%$ confidence interval. The parameters varied are the number of superpixels during segmentation (``sparse,'' ``medium,'' and ``dense''), and the ball query radius. The plots show that increasing the edges density of the graphs increases test prediction accuracy up to a point. Note that the scale for the y-axis is different for each plot.}
        	\label{fig:edges}
        \end{figure}

In this Section, we examine the influence of node and edge density of the graph representations on model prediction accuracy. We controlled graph nodal density by prescribing ``sparse,'' ``medium,'' and ``dense'' levels of segmentations to the high-resolution images (see Fig. \ref{fig:ML}a). Details on the distribution of node densities for each sub-dataset as well as visualizations of graphs from each node density are presented in Appendix \ref{appen:node_den}. We controlled edge density by defining different query radii of $r=0.2w,0.3w,0.4w$, where $w$ is the width of the column. Note that for a fixed ball query radius, simply increasing the nodal density will increase the edge density of the graph. 
In Fig. \ref{fig:edges}, we illustrate the effect of graph edge density on prediction accuracy. Again, we note that each data point contains the mean $\pm 95\%$ confidence interval (CI) of $10$ separately trained models with the same architecture but different initializations. In all cases, the ML model is trained with an augmented dataset representing $20,000$ simulation results. As demonstrated in Fig. \ref{fig:edges}, regardless of neighborhood size, the prediction accuracy increases when edge density increases up to a maximum prediction accuracy where adding in additional edges does not improve performance. This result may be due to the fact that after each message passing layer, the subsequent message passing layers will contain information from the next hop (neighbors of neighbors), thus higher edge density is redundant \citep{gnn} since no additional information about the geometry will be added. With our chosen model architecture of $4$ Pointnet++ layers, each node will have information about its $4$-hop neighbors. 
The average number of edges for which the the highest prediction accuracies are achieved is $2046$ edges for sub-dataset 1 (``medium'' node density with query radius of $r=0.4w$), $3815$ edges for sub-dataset 2 (``dense'' node density with query radius of $r=0.3w$), and  $7479$ edges for sub-dataset 3 (``dense'' node density with query radius of $r=0.3w$). Again, the maximum prediction accuracy is shown in Table \ref{tab:aug} column $4$. 

\subsection{Ball Query Graph Representation Out-Performs the Region Adjacency Graph and Perfect Structure Representation}
\label{sec:res_rag}

\begin{figure}[ht]
        	\centering
        	\includegraphics[width=\textwidth,keepaspectratio]{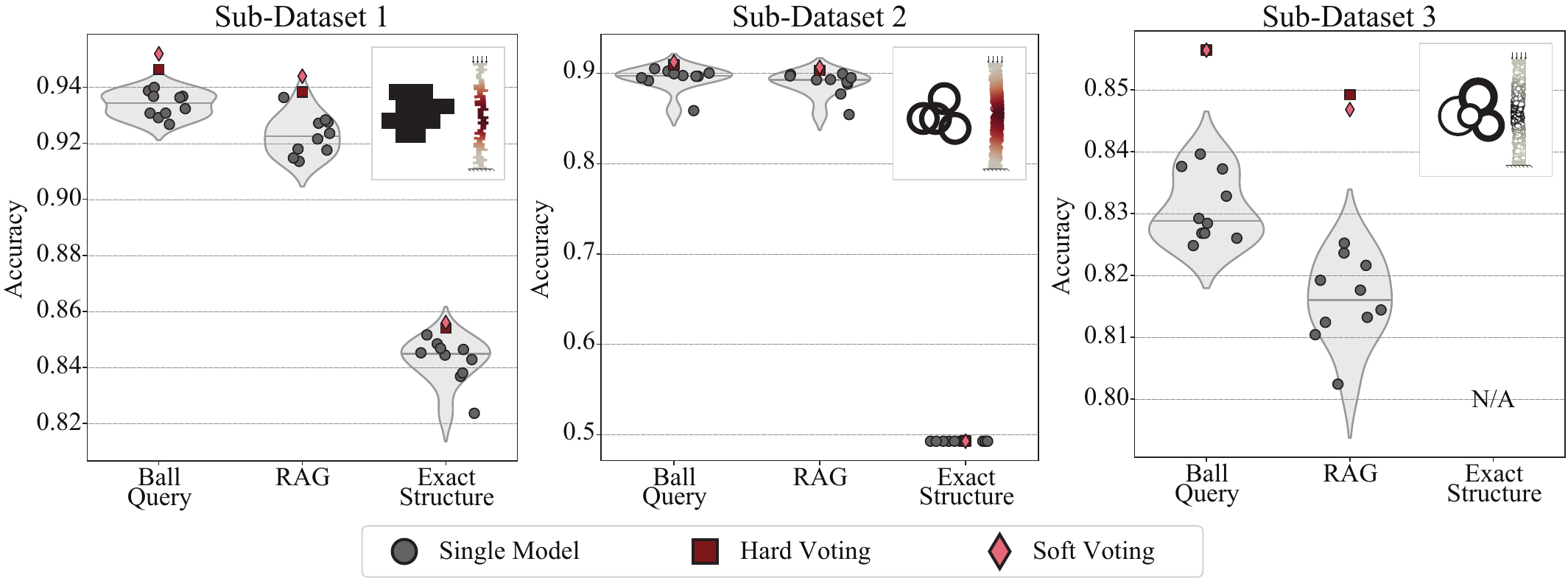}
        	\caption{Test accuracy with respect to graph structure representation method for an augmented training dataset of $20,000$ simulations with node density and ball radius informed by the parameter sweep in Section \ref{sec:meth_representation}. In each plot, the black circle marker shows the accuracy of an individual ML model. The red square and pink diamond show aggregated accuracy from hard voting and soft voting respectively. For each dataset, we show results from input graphs created through ball query, RAG, and exact representation. Note that the scale for the y-axis are  different for each plot, and that for sub-dataset 3 there is no ``exact structure'' representation. }
        	\label{fig:jitter}
\end{figure}
        
Here, we investigate three approaches to column geometry representation and observe the effect of each approach on pipeline prediction accuracy. We note that aside from the different input graphs for training, validating, and testing phases, everything else in the pipeline (e.g. the ML architecture and the $10$ different weight initializations) remains the same. As stated in Section \ref{sec:meth_representation}, we investigate ball query representation as the primary approach, and region adjacency graph (RAG) representation and ``exact structure'' representation as two potential alternatives. Figure \ref{fig:jitter} illustrates the test prediction accuracies of each input representation type. For sub-dataset 1 and sub-dataset 3, RAG representations perform slightly worse than the ball query approach. For sub-dataset 2, RAG performs comparably to ball query, but is still marginally worse. We note that RAG representations can be seen as a special case of the ball query algorithm where only neighboring superpixels form edges (i.e., query radius $r$ matches the superpixel size), leading it to perform similarly to graphs constructed using the ball query approach. Critically, we note that despite no loss of geometric information, the ``exact" representation approach performs worse than both ball query and RAG for both sub-dataset 1 and sub-dataset 2 (for sub-dataset 3 there is no ``exact'' structure formulation available due to the trimming step of geometry generation). The difference in performance is particularly striking for sub-dataset 2, where the ``exact'' representation prediction accuracy is $\sim50\%$ for all model initializations, indicating that the ML model does not perform better than random guessing. We believe that this was because PointNet++ was initially developed to deal with point cloud datasets such as ShapeNet \citep{shapenet, pointnet}, so point cloud analogous geometric representations work best with this pipeline. We note that there could potentially be other ML model architectures where exact geometry representations would perform better, but exploration of these architectures is outside the scope of this paper. 

\subsection{Ensemble Learning Enhances Prediction Accuracy}
\label{sec:res_vote}
        
\begin{table}[ht]
\label{tab:ensemble}
\centering
\caption{Comparison of test prediction accuracies for individual model mean, best individual model, hard voting, and soft voting. This table reflects the $10$ ball query based models for each sub-dataset shown in Fig. \ref{fig:jitter}.}
\begin{tabular}{|l||ccc|}
\hline
&\textbf{Sub-Dataset 1} & \textbf{Sub-Dataset 2} & \textbf{Sub-Dataset 3} \\
\hline
\textbf{Mean of 10 Models} & 0.934 & 0.894 & 0.831 \\
\textbf{Best Individual Model} & 0.937 & 0.905 &  0.839 \\
\textbf{Hard Voting} & 0.946 & 0.910 & 0.856 \\
\textbf{Soft Voting} & 0.952 & 0.913 & 0.856 \\
\hline
\end{tabular}
\end{table}

For each sub-dataset, we independently train $10$ ML models with the same hyperparameters but different weight initializations. Therefore, we can investigate not only the average and range of model accuracy, but also the efficacy of applying ensemble learning methods to leverage all independently trained models to boost performance. Specifically, we employ both hard voting and soft voting to aggregate our predictions from the available ML models (see Section \ref{sec:meth_ensemble} for details on hard and soft voting). The plots in Fig. \ref{fig:jitter} demonstrate the effectiveness of both voting methods for increasing the prediction accuracy for column representations  produced from our proposed metamodeling pipeline outlined in Section \ref{sec:meth_representation}, particularly for sub-dataset 3. In all cases, the final predictions obtained via voting lead to slightly better prediction accuracy than the best individual accuracy from the $10$ different model initializations (see Table \ref{tab:ensemble}). For sub-dataset 1, soft voting performed better than hard voting and lead to an accuracy increase of $\sim0.15\%$. For sub-dataset 2, soft voting performed better than hard voting and lead of an accuracy increase of $\sim0.076\%$. For sub-dataset 3, soft voting and hard voting performed similarly and lead to an accuracy increase of $\sim1.7\%$. The efficacy of ensemble voting with our classifier suggests that differently initialized models potentially learn different geometric features and combining these trained models leads to overall variance reduction \citep{bagging,ensemblednn,ensemble,boosting}. \changes{The visualizations of the important edges with GNNExplainer \citep{gnnexplainer} for each sub-dataset in Appendix \ref{appen:vis} further support the notion that different geometric features are learned in different initializations. However, due to the different geometrical features that are highlighted depending on model initialization, it is not clear if important insights on the geometry can be inferred from GNNExplainer.}

\subsection{\changes{Ensemble Learning Displays Well-Calibrated Confidence}}
\label{sec:calibration}
\begin{figure}[h]
        	\centering
        	\includegraphics[width=\textwidth,keepaspectratio]{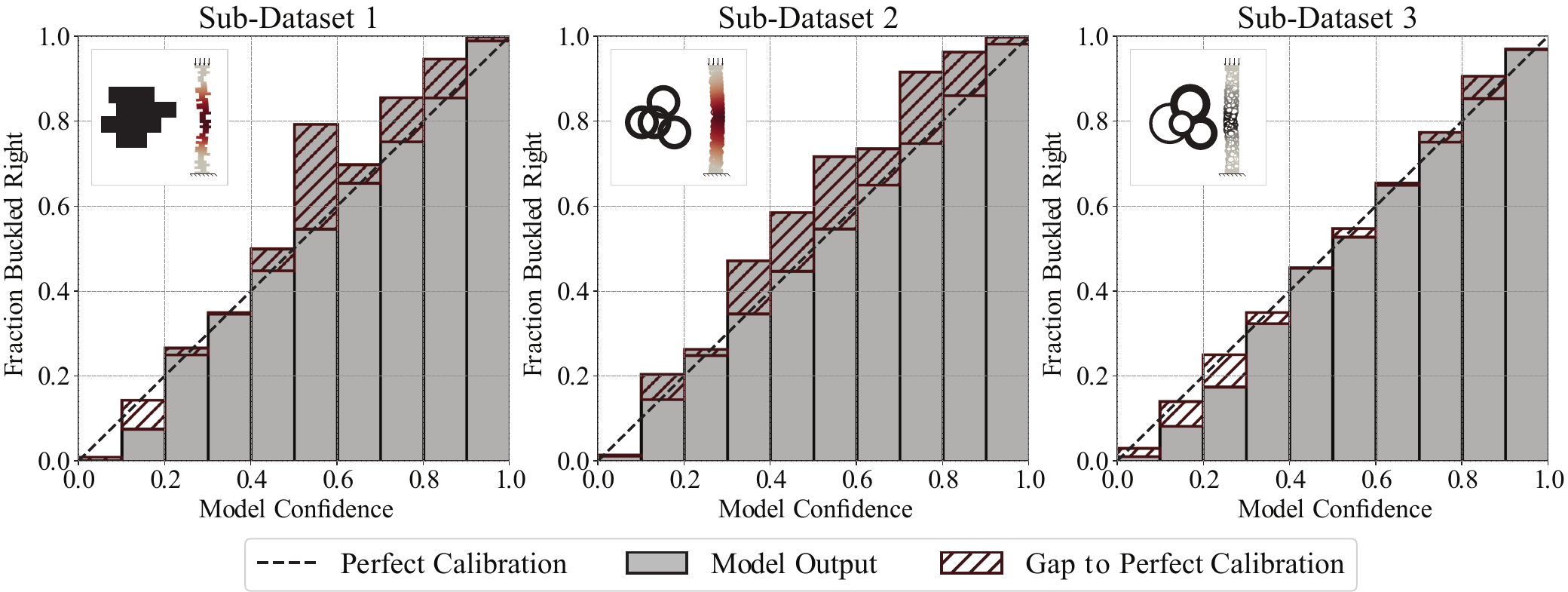}
        	\caption{\changes{Reliability Curves for the best performing ensemble of $10$ models. Models are trained on  $r = 0.4w$ with ``medium'' node density for sub-dataset 1, and ``dense" node density with $r = 0.3w$ for sub-dataset 2 and sub-dataset 3. A perfectly calibrated model would be a perfect diagonal. In this plot, the gap between observed and perfect calibration is noted with the hatched pattern.}}
        	\label{fig:confidence}
\end{figure}

\begin{table}[h]
\centering
\caption{\changes{Comparison of Expected Calibrated Error ($ECE$, see eqn. \ref{eqn:ECE}) and Maximum Calibrated Error ($MCE$, see eqn. \ref{eqn:MCE}) for ensemble of ML models. A perfectly calibrated model would have $ECE$ and $MCE$ of zero.}}
\label{tab:ECEMCE}
\begin{tabular}{|c||ccc|}
\hline
\multicolumn{1}{|l||}{} & \textbf{Sub-Dataset 1} & \textbf{Sub-Dataset 2} & \textbf{Sub-Dataset 3} \\
\hline
\textbf{$ECE$}                  & 0.024        & 0.041         & 0.023         \\
\textbf{$MCE$}                  & 0.246        & 0.169         & 0.076\\
\hline
\end{tabular}
\end{table}

\changes{In addition to evaluating the accuracy of our ML models, we are also interested in evaluating the confidence of our ensemble of predictions. As referred to in Section \ref{sec:meth_ensemble}, reliability diagrams are a useful visual tool to evaluate model calibration. To construct our reliability diagrams, we first separate the model predictions into bins according to the model confidence that the input column is in class $1$ (i.e. buckles right), and then count the actual number of class $1$ labels in the bin. The number of class $1$ labels is then normalized with the number of samples in each bin. Figure \ref{fig:confidence} shows the reliability diagrams for the ensemble of $10$ best-performing models defined in Section \ref{sec:res_rep} that are trained on ``medium'' node density with $r = 0.4w$ for sub-dataset 1, and ``dense'' node density with $r = 0.3w$ for both sub-dataset 2 and sub-dataset 3. In Table \ref{tab:ECEMCE}, we report the $ECE$ and $MCE$ for each ensemble. Recall that for model confidences to be well-calibrated, the reliability diagram must be close to the diagonal (i.e., the black dotted line in Fig. \ref{fig:confidence}). As such, perfectly-calibrated models will also have $ECE$ and $MCE$ values that are close to zero. Visually, the reliability diagrams in Fig. \ref{fig:confidence} shows that our ML models are slightly underconfident (lower confidence compared to fraction of class $1$), but overall still exhibit reliable behavior. Quantitatively, the $ECE$ and $MCE$ for each ML models is relatively low, especially in the case of sub-dataset 3.  In Appendix \ref{sec:comparison_uncertain}, we further investigate potential differences in the deformation profile between columns from different model confidence levels.} 

\section{Conclusion}
\label{Conclusion}

In this paper, we introduce the Asymmetric Buckling Columns (ABC) dataset, a mechanics specific classification dataset of spatially heterogenous columns under compression where local geometric structure influences the global emergent buckling direction. The dataset consists of $3$ sub-datasets, each constructed from basic geometric features with varying degrees of complexity. 
Our goal in defining these complex geometries was to create a dataset where treating each column as an image-like array would not be the obvious choice for data representation. \changes{Then, to explore alternatives to ``image-like'' data representations while keeping the benefits of inductive biases from the convolution operation, we proposed a metamodeling pipleline that is broadly applicable to any structure with complex geometric features, and tested its performance on the ABC dataset.} The proposed pipeline first converts the columns into spatial graphs via image segmentation, then employs a ML architecture comprised of PointNet++ spatial graph convolution layers \citep{pointnet} to classify each column based on its buckling direction. Within the scope of our proposed pipeline, we also investigated strategies to increase prediction accuracy through data augmentation, alternative spatial graph formulations, and ensemble learning. We showed that our data augmentation scheme improved the prediction accuracy comparably to increasing the training set size by a factor of $4$. We also demonstrated that choosing a suitable edge density to represent our structure is key to obtaining the best prediction accuracy. Additionally, we improved the overall prediction accuracy by applying voting methods to our $10$ separately trained models. \changes{Finally, we showed that our proposed ensemble of metamodels is well-calibrated}.

Looking forward, we hope that this work will serve as a platform for other researchers to explore alternative data representations for ML pipelines in mechanics, especially in problems where geometry plays an important role in the emergent behavior of the system. To facilitate this, we have released the ABC dataset, the code to generate the dataset, and the code to implement our proposed ML pipeline all under open source licenses.
The availability of these resources will allow others to either apply our ML pipeline to alternative problems, or benchmark alternative methods on the same dataset.
Beyond the scope of the classification problem presented in this paper, it would also be interesting to see how different structure representation strategies and GNN architectures perform for predicting full-field QoIs for various problems in mechanics. Specifically, while there are reported successes in predicting full-field QoIs in a variety of mechanics problems \citep{DLfullfield,cGANfullfield,bayesianfullfield}, and examples of applying GNNs to predict full-field QoIs from physics-based simulations \citep{gnnmesh,gnnparticle}, there are limited examples in the literature that predict full-field QoIs of truly emergent mechanical behavior from unstructured and non-gridlike data. 
Similarly, substantial future work is needed in curating and disseminating benchmark datasets of complex structures that are experimentally tested \citep{bear, breadbuckle} rather than simulated. Due to the time and cost for experiments, methods to combine experimental data and computational data through methods such as transfer learning should also be explored \citep{beartransfer, transfermetamodel}.
Another interesting direction for future work is in the interpretability of GNNs especially when applied to problems concerning solid mechanics. While GNN interpretibility for problems in computer vision typically focuses on qualitative visualizations \citep{gnnexplainer}, there have also been important recent advances in interpreting learned physical laws as simple symbolic equations with GNNs on discrete n-body problems \citep{SymbolicGNN1,SymbolicGNN2}. It would be interesting to see if similar methods could be applied to problems in solid mechanics to potentially interpret the complex relations between structural behavior and domain geometry. To this end, we view the creation of the ABC dataset and our initial pipeline as a starting point for substantial future work at the intersection of machine learning and mechanics.

\section{Additional Information}
\label{additionalinfo}

The ABC dataset is available through the OpenBU
Institutional Repository at \url{https://open.bu.edu/handle/2144/43730} \citep{ABC}. All code used to generate the the domain geometry and labels is available at \url{https://github.com/pprachas/ABC_dataset}. All code for implementation of the metamodeling pipeline from spatial graph generation to PointNet++ layers is also available at \url{https://github.com/pprachas/ABC_dataset}.

\section{Declaration of competing interest}
The authors declare that they have no known competing financial interests or personal relationships that could have appeared to influence the work reported in this paper.

\section{Acknowledgements}
We would like to thank the staff of the Boston University Research Computing Services and the OpenBU Institutional Repository (in particular Eleni Castro) for their invaluable assistance with generating and disseminating the Asymmetric Buckling Columns (ABC) Dataset. This work was made possible through start up funds from the Boston University Department of Mechanical Engineering, the David R. Dalton Career Development Professorship, the Hariri Institute Junior Faculty Fellowship, the Haythornthwaite Research Initiation Grant, and the National Science Foundation grant CMMI-2127864.

\newpage

\appendix

\section{Node density}
\label{appen:node_den}

\begin{figure}[h]
        	\centering
        	\includegraphics[width=\textwidth,keepaspectratio]{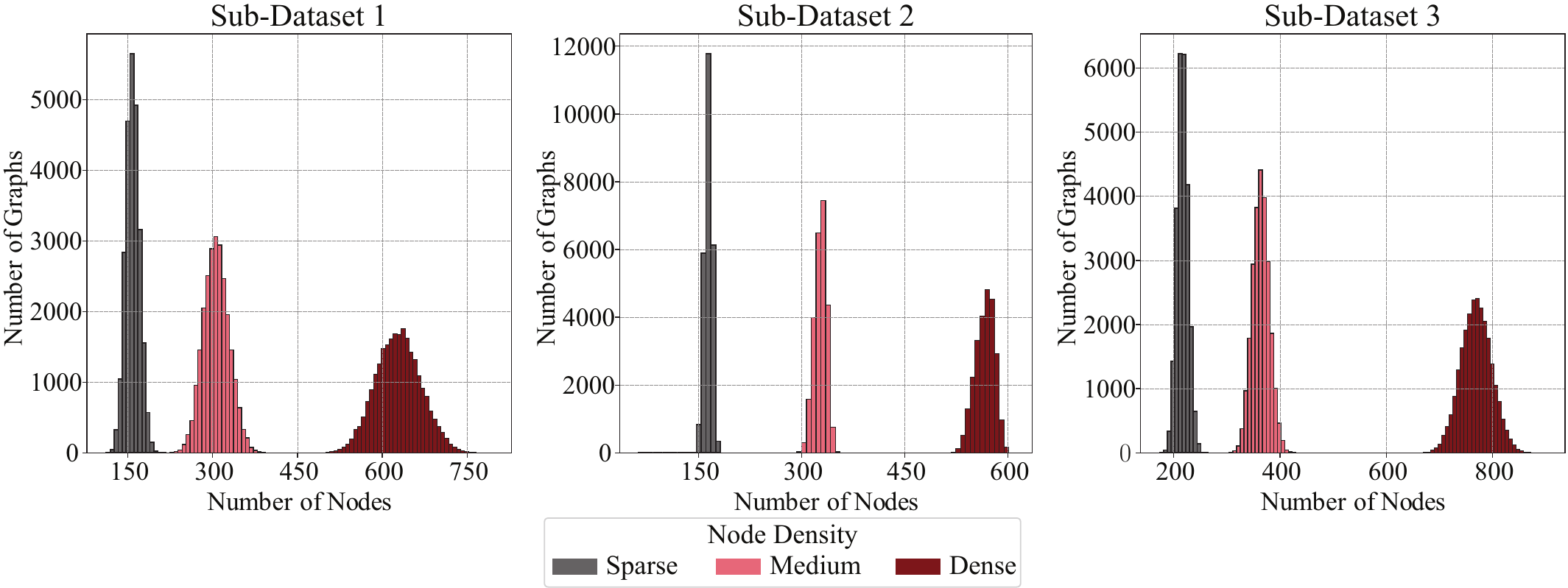}
        	\caption{Histogram of graph node densities from our dataset. Each plot contains histograms of sparse, medium, and dense node densities for each sub-dataset.}
        	\label{fig:node_den}
\end{figure}

\begin{figure}[h]
        	\centering
        	\includegraphics[width=\textwidth,keepaspectratio]{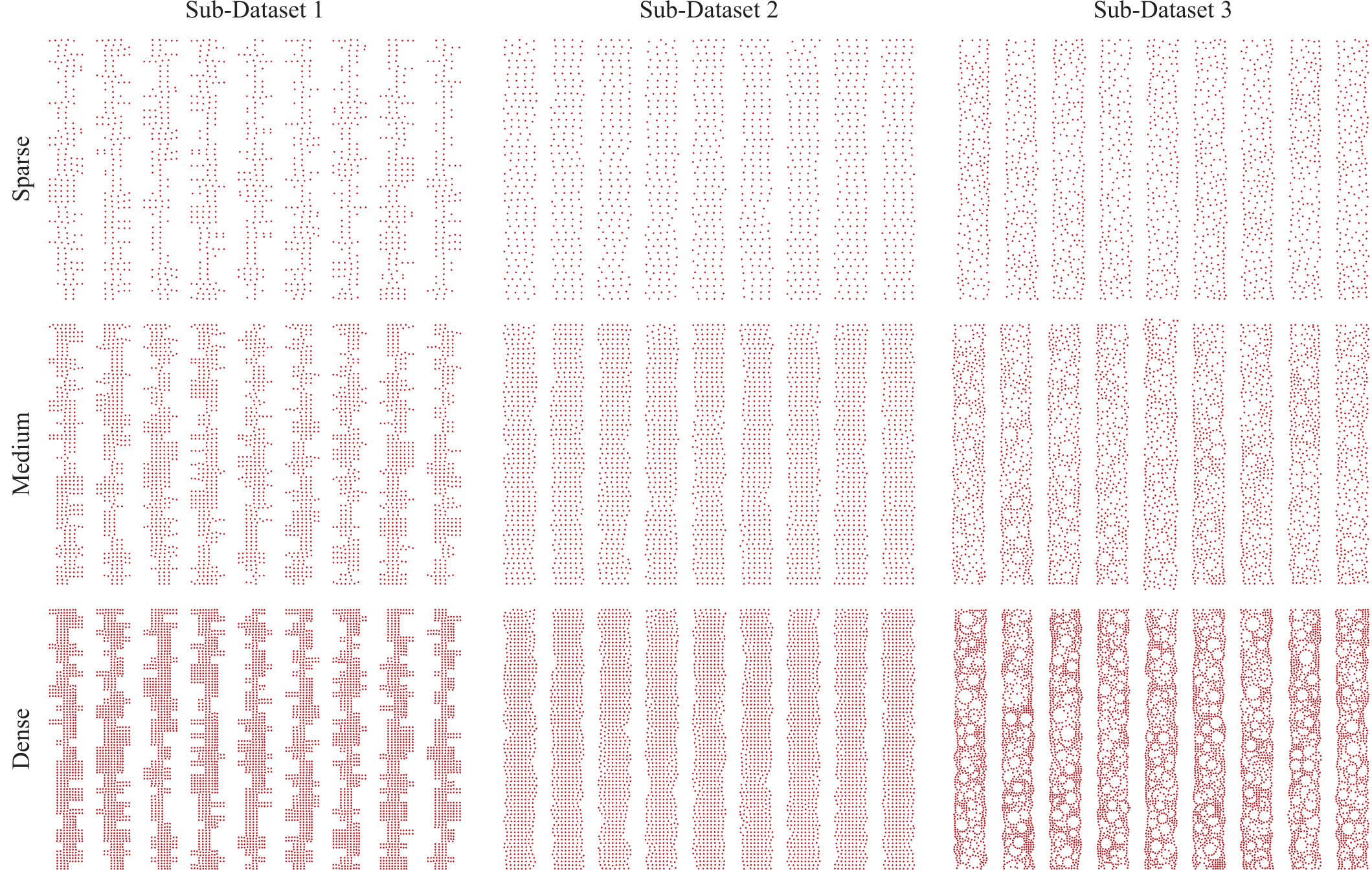}
        	\caption{Comparisons of node densities of the ABC columns obtained from our metamodeling pipeline for each sub-dataset. The structures in each row are ordered from sparse (top) to dense (bottom).}
        	\label{fig:node_denvis}
\end{figure}

Figure \ref{fig:node_den} illustrates the histograms of graph node densities from using the metamodeling pipeline proposed in Section \ref{sec:meth_representation}. We note that the node densities are not fixed for each dataset since we controlled the number of superpixels during SLIC segmentation, and superpixels not associated with the structure (the black regions in Fig. \ref{fig:ML}) are discarded. Fig. \ref{fig:node_denvis} gives a visual comparison between the different nodal densities for each sub-dataset.

\section{Visualization of ML architecture with different initializations}
\label{appen:vis}

\begin{figure}[h]
        	\centering
        	\includegraphics[width=.875\textwidth,keepaspectratio]{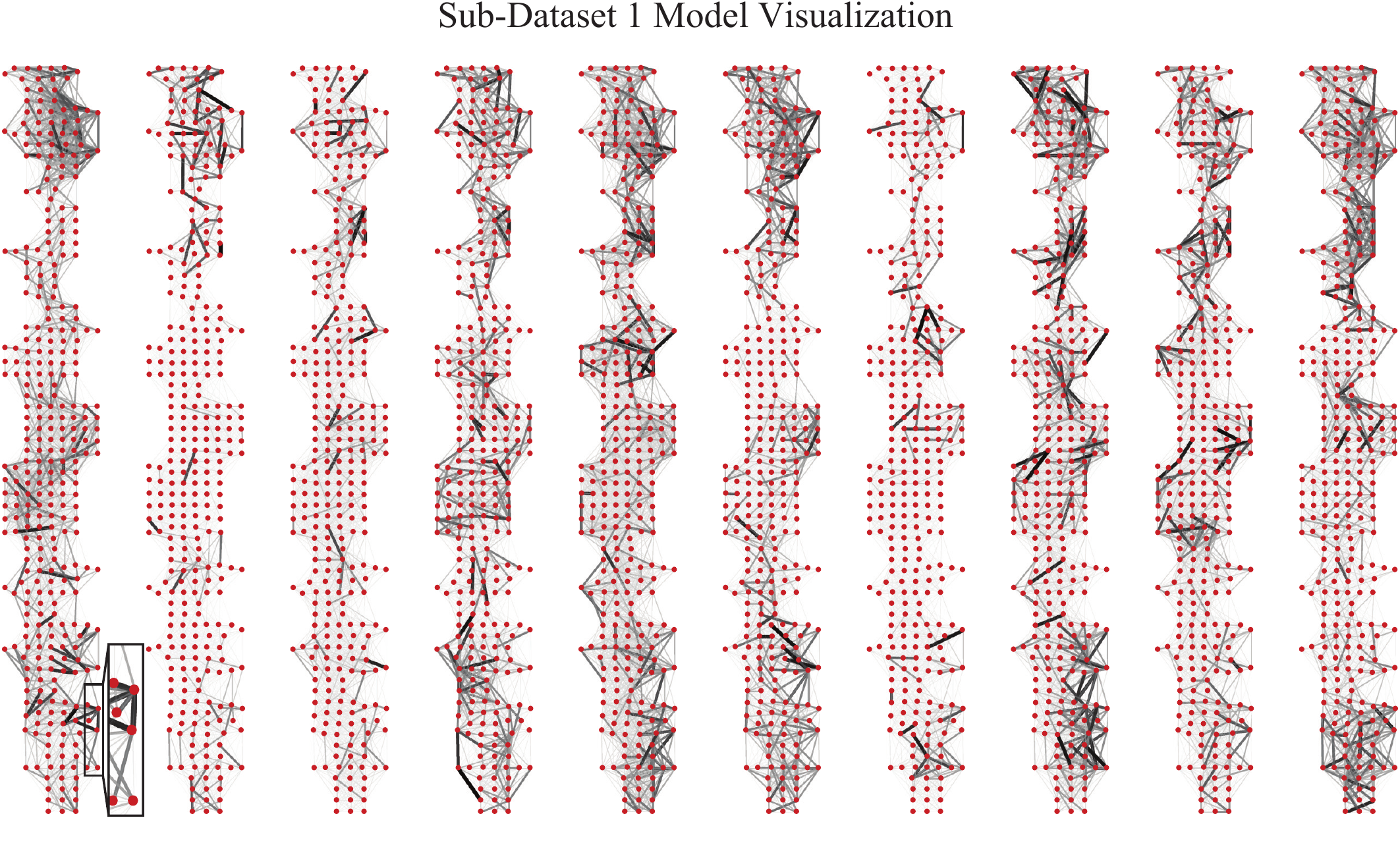}
        	\caption{Visualization of the $10$ different initializations of the trained GNN model for sub-dataset 1 with medium node density. In this visualization, created with GNNExplainer \citep{gnnexplainer}, nodes are represented as red markers, all nodes within ball radius $40$ are connected by edges, and edges that are likely to be influential to label predictions are darker and thicker (see zoom of the first structure).}
        	\label{fig:vis1}
\end{figure}

\begin{figure}[!htb]
        	\centering
        	\includegraphics[width=.875\textwidth,keepaspectratio]{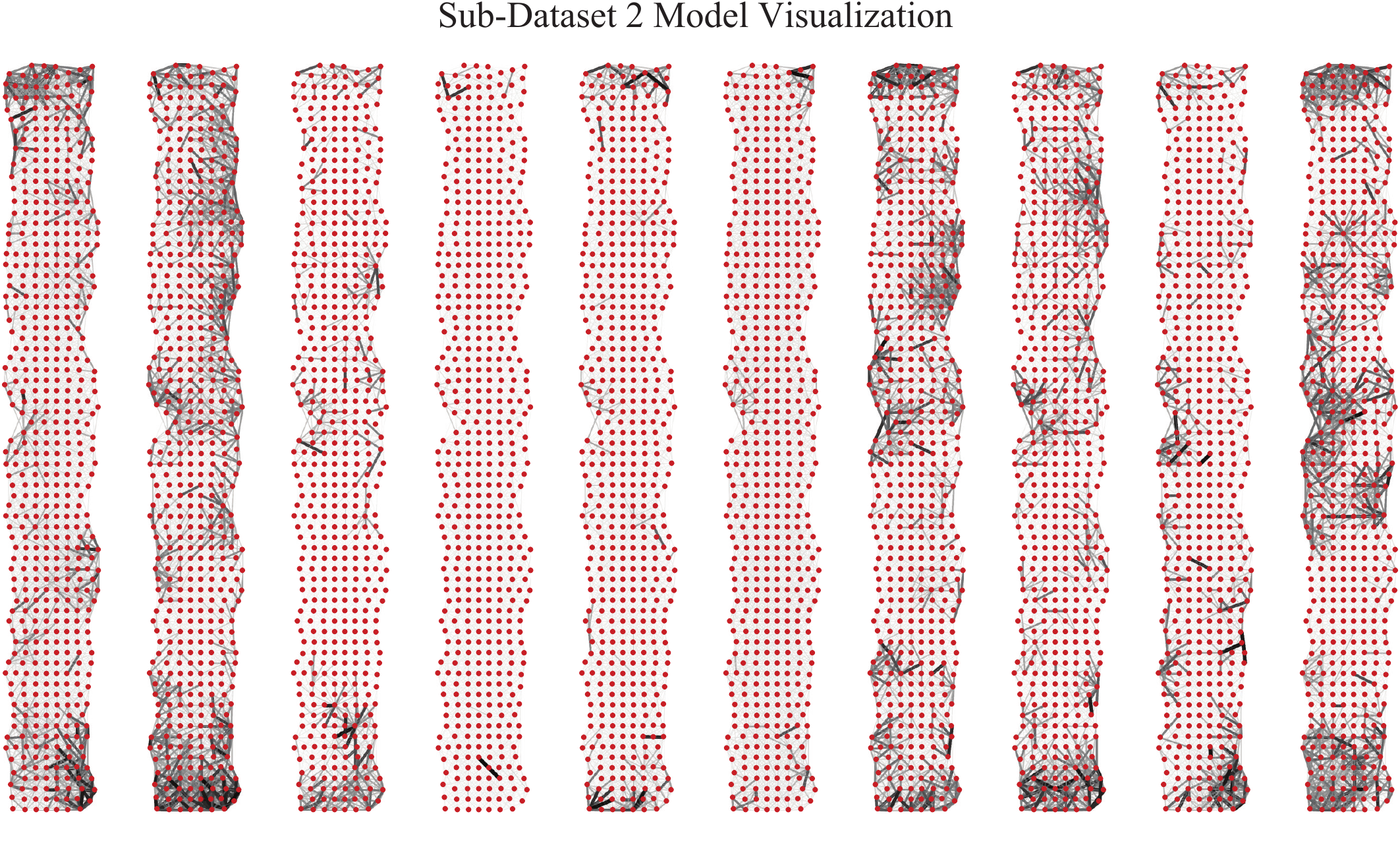}
        	\caption{Visualization of the $10$ different initializations of the trained GNN model for sub-dataset 2 with dense nodal density. In this visualization, created with GNNExplainer \citep{gnnexplainer}, nodes are represented as red markers, all nodes within ball radius $30$ are connected by edges, and edges that are likely to be influential to label predictions are darker and thicker.}
        	\label{fig:vis2}
\end{figure}

\begin{figure}[!htb]
        	\centering
        	\includegraphics[width=.875\textwidth,keepaspectratio]{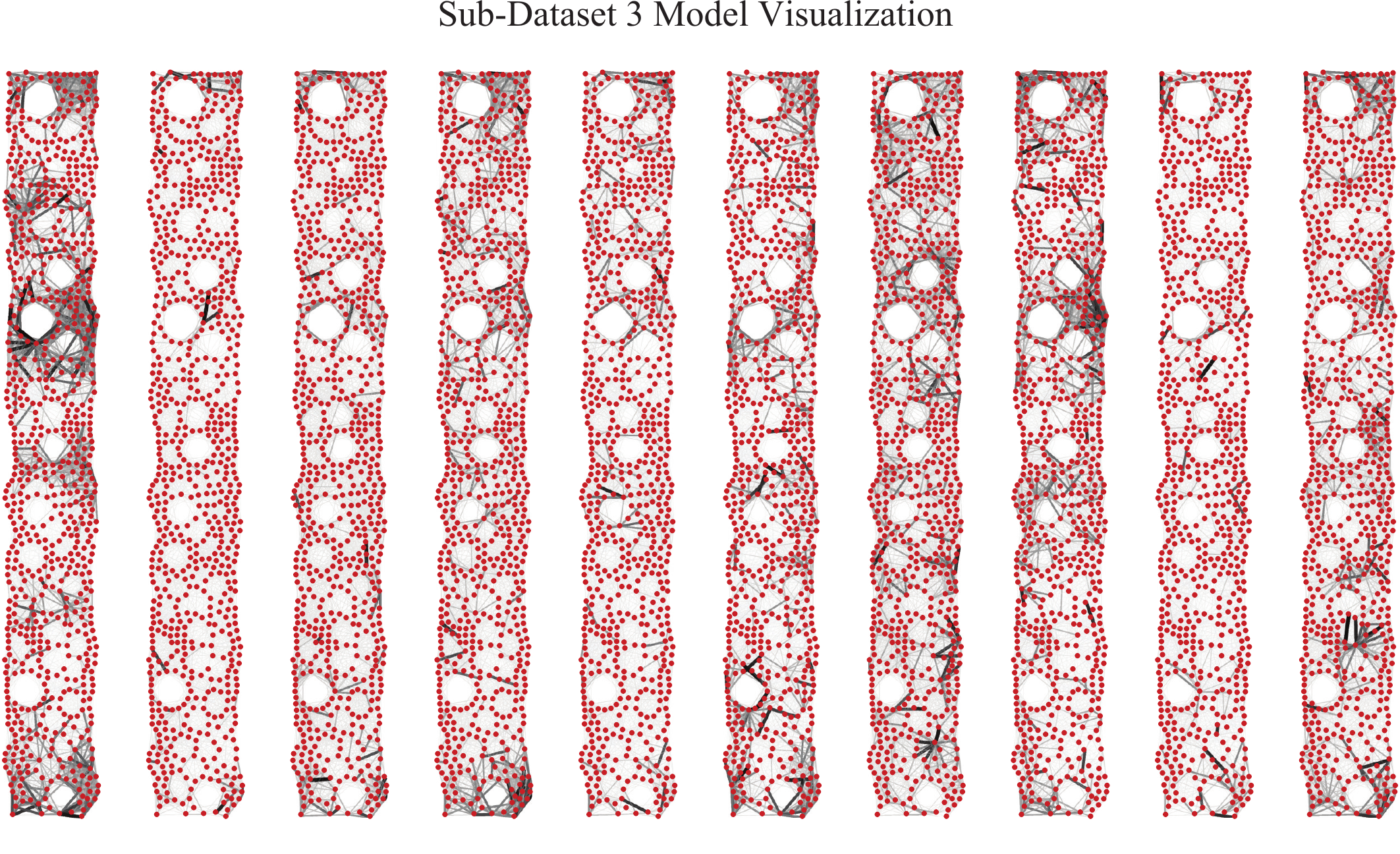}
        	\caption{Visualization of the $10$ different initializations of the trained GNN model for sub-dataset 3 with dense nodal density. In this visualization, created with GNNExplainer \citep{gnnexplainer}, nodes are represented as red markers, all nodes within ball radius $30$ are connected by edges, and edges that are likely to be influential to label predictions are darker and thicker.}
        	\label{fig:vis3}
\end{figure}

In this Appendix, we visualize our trained ML models with GNNexplainer \citep{gnnexplainer}. In brief, GNNExplainer interprets the decisions of a GNN model by identifying the sub-graph $G_S$ of the input graph that is the most important to the prediction of the output label. We note that in our case of graph classification, $G_S$ does not have to be connected. For large graphs, brute-force testing of all possible sub-graphs is computationally intractable. The GNNExplainer algorithm works around this constraint by determining the probability that each edge is part of sub-graph $G_S$ \citep{gnnexplainer}. As such, the output of GNNExplainer is a weighted edge-mask of a given graph structure that determines the probability that each edge is a part of $G_S$ for a given trained GNN model. In Fig. \ref{fig:vis1}-\ref{fig:vis3}, we visualize the edge-masks for each of our trained ML models applied to identical samples. For each example, the underlying graph structure is identical (created from our metamodeling pipeline), and GNNExplainer is applied to the $10$ different ML model initializations. Edges with a higher probability of being part of $G_S$ are darker and thicker (see in particular the Fig.\ref{fig:vis1} zoom of the first structure). We note that while the model visualization highlights interesting features, in all cases, the features highlighted by GNNExplainer are not consistent across models. This observation leads us to believe that each ML model is learning different features, and therefore prompted us to investigate ensemble methods as a means to increase prediction accuracy (see Section \ref{sec:meth_ensemble}). 
We also note that GNNExplainer outputs a feature mask that shows the importance of each node. Similar to the edge masks, we observed that each initialization led to a different importance for each nodal feature, further supporting the notion that the models from each initialization learn different features. \changes{On the other hand, since $G_S$ depends on model initializations, it is not clear if any insights on mechanistically important features can be obtained through GNNExplainer directly.}


\section{\changes{Comparison between Deformed Profiles of High and Low Confidence Predictions}}
\label{sec:comparison_uncertain}

\begin{figure}[h]
        	\centering
        	\includegraphics[width=\textwidth,keepaspectratio]{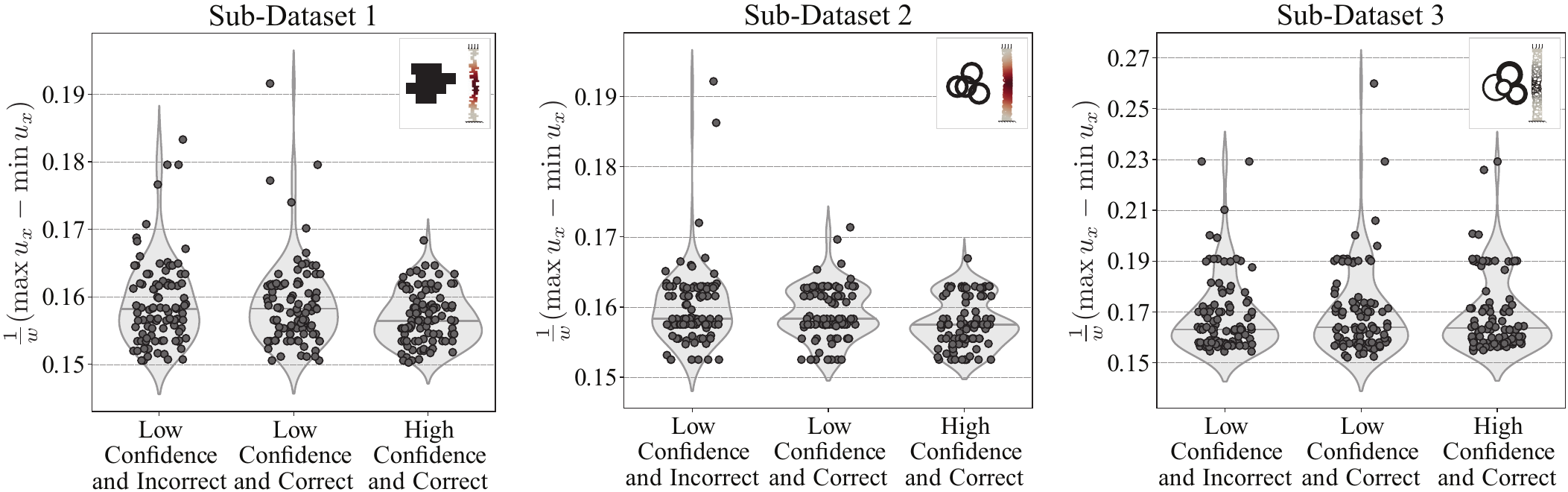}
        	\caption{\changes{Difference of maximum normalized x-displacements for each category, with $100$ columns in each category. The proposed metric roughly measures the potential presence of higher order buckling modes, which is a source of uncertainty. We note that our stopping criterion for our simulations is defined as $\max | u_x | > 0.15w$.}}
        	\label{fig:dispdiff}
\end{figure}

\begin{figure}[p]
        	\centering
        	\includegraphics[width=\textwidth,keepaspectratio]{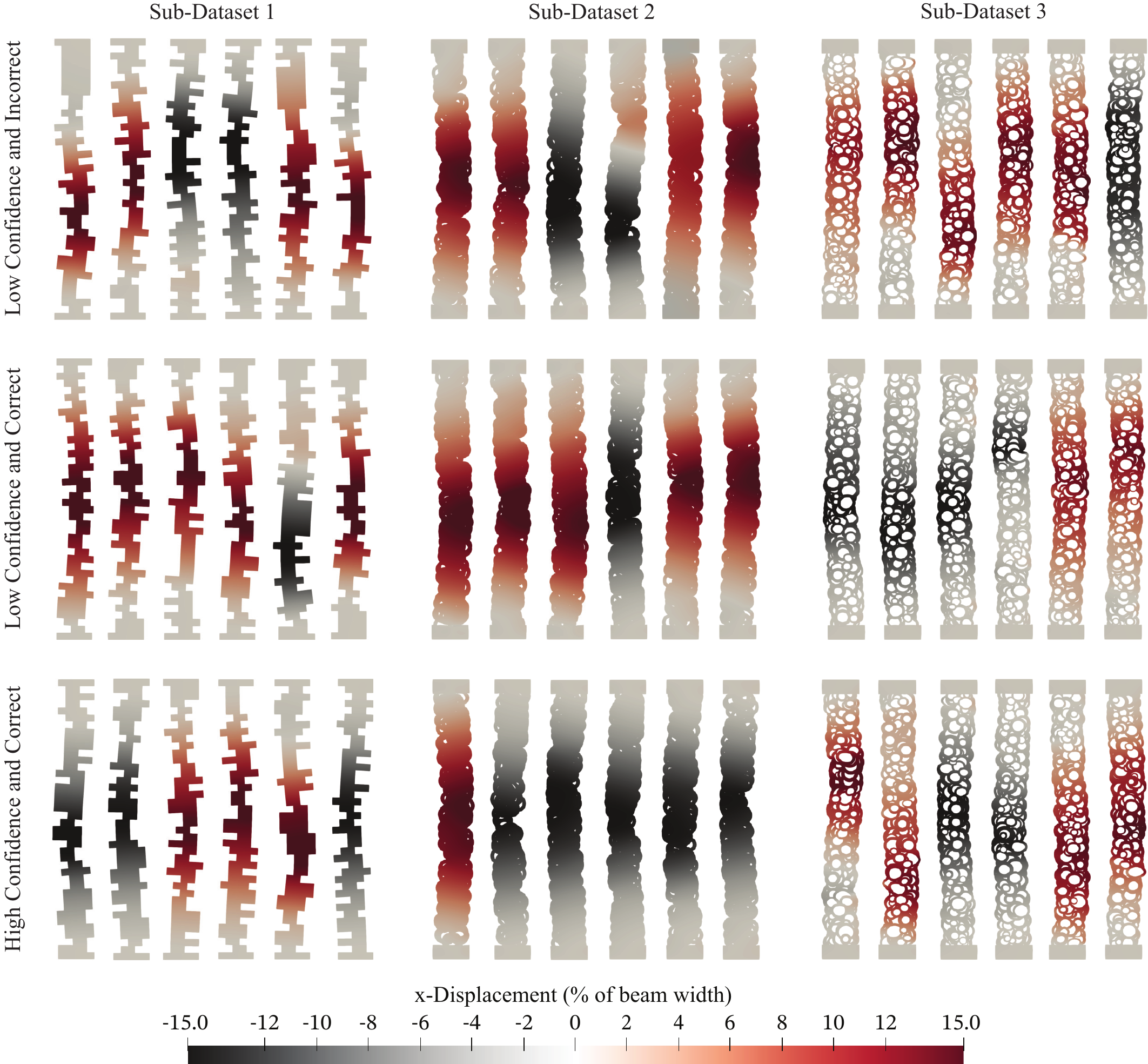}
        	\caption{\changes{Examples of deformed columns for high and low confidence predictions. Structures are arranged according to confidence levels (with low being ~$50 \%$ probability of being in either class and high being $100\%$ in the correct class) and accuracy of prediction.}}
        	\label{fig:deformed}
\end{figure}

\begin{figure}[p]
        	\centering
        	\includegraphics[width=\textwidth,keepaspectratio]{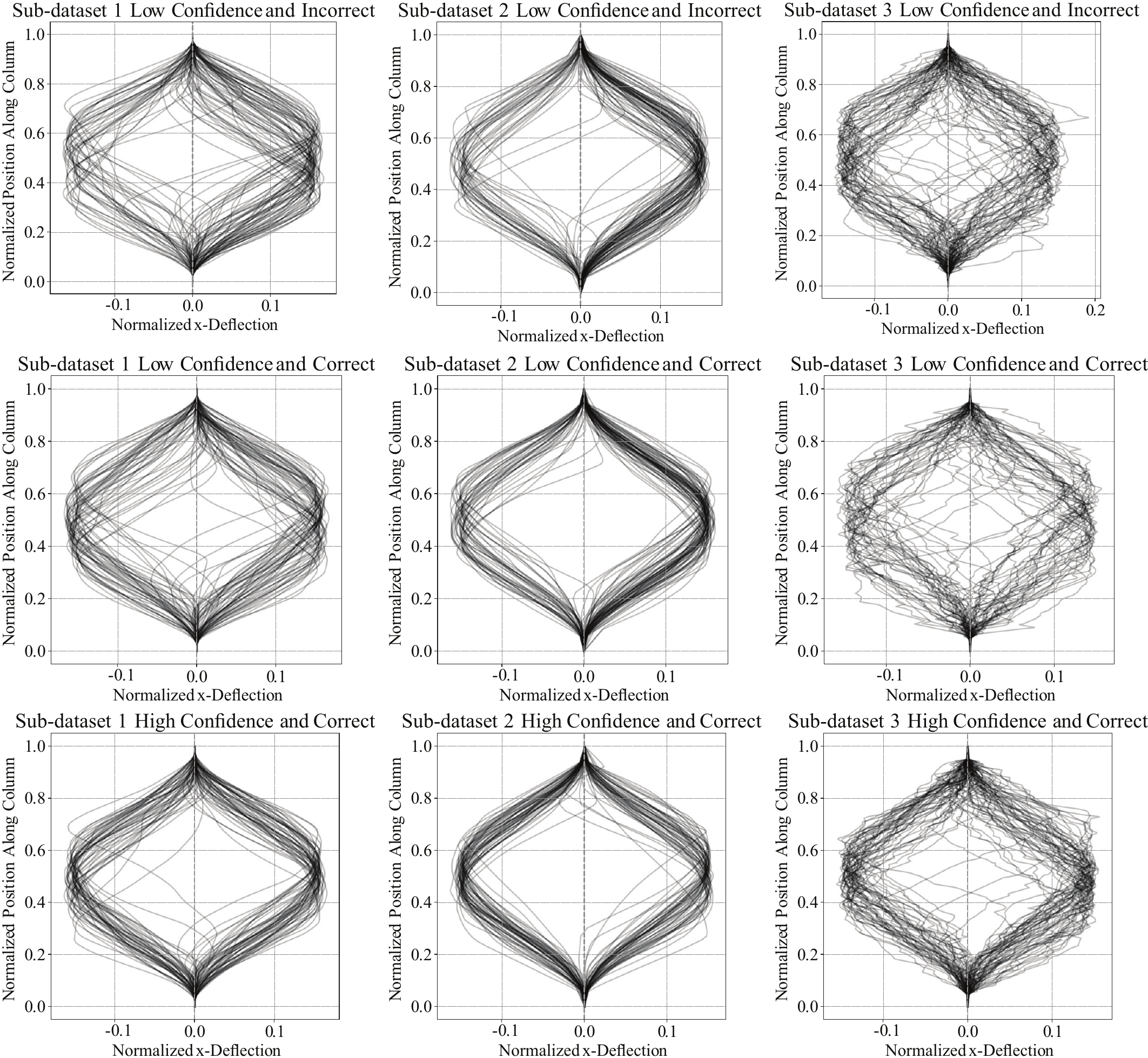}
        	\caption{\changes{Centerline displacements of $100$ columns in each category of confidence and accuracy. Plots are arranged according to confidence levels (with low being ~$50 \%$ probability of being in either class and high being $100\%$ in the correct class) and accuracy of prediction. Note that x-axis and y-axis are scaled differently for ease of visualization, and thus the illustrated deflections are not to scale.}}
        	\label{fig:centerline}
\end{figure}

\changes{After confirming that our metamodel ensembles are well-calibrated, we investigate potential patterns across three cases: (1) low confidence with incorrect predictions (``Low Confidence and Incorrect''), (2) low confidence with correct predictions (``Low Confidence and Correct''), and (3) high confidence with correct predictions (``High Confidence and Correct''). To this end, we visualize randomly selected examples from each case in Fig. \ref{fig:deformed}, we visualize overlaid column centerline displacements from each case in Fig. \ref{fig:centerline}, and we visualize the range of x-displacement present in each column in Fig. \ref{fig:dispdiff}. Across all examples, there is little overt visible difference between cases. However, in sub-dataset 1 and sub-dataset 2, some columns that correspond to low confidence predictions exhibits signs of higher modes of buckling (e.g., centerline x-deflections cross the $y=0$). However, this qualitative difference between cases is not observed in subdataset-3, perhaps due to the fact that local buckling of trimmed rings in subdataset-3 lead to more complex deformation profiles as shown in Fig.\ref{fig:deformed} and Fig.\ref{fig:centerline}.}

\changes{In Fig.\ref{fig:dispdiff}, we plot the range of x-displacements for each column for $100$ columns per group to further attempt to visualize a trend. Specifically, Fig.\ref{fig:dispdiff} shows the distribution of the difference in maximum and minimum x-displacement that is normalized by the column width ($(1/w)(\max u_x - \min u_x)$). Note that the stopping criterion for our simulations is a maximum absolute x-displacement greater than $15\%$ of the column width ($\max u_x > 0.15w$). Again, while there is some distinctions in the differences in deflection between high confidence and low confidence predictions in the case of sub-dataset 1 and sub-dataset 2,  there is no real separation in the case of sub-dataset 3. 
This lack of clear distinction between the deformation profiles for each case might be due to the fact that uncertainty in model predictions comes from a combination of sources. First, the model performance plots shown in Fig. \ref{fig:aug} clearly indicate that imperfect model performance is in part due to small training set size. Essentially, lack of coverage of the massive input parameter space is a source of model uncertainty. Second, as shown in Fig. \ref{fig:ML}a, each structure geometry is captured approximately rather than exactly, which could potentially contribute to uncertainty in prediction. Finally, and perhaps most interestingly, some of the qualitative results shown in Fig. \ref{fig:centerline} indicate that a subset of the columns in this dataset could be behaving a way that is ``mechanically distinct'' where the buckled column deforms in a manner that is more dissimilar from typical first mode buckling of a symmetrically loaded fixed-fixed column without geometric heterogeneity. We consider this third potential source of uncertainty a particularly interesting avenue for further study.}

\FloatBarrier 
\newpage
\bibliographystyle{plain}
\bibliography{main}

\end{document}